\pdfoutput=1 

\documentclass[sigconf]{acmart}

\usepackage{textcomp}
\usepackage{amssymb}
\usepackage{pifont}
\usepackage{subfigure}
\usepackage{tabularx}
\usepackage{amsmath,amsthm}
\usepackage{graphicx}

\usepackage{enumitem}
\usepackage{array,makecell}
\usepackage[T1]{fontenc}
\usepackage{algorithmic}
\usepackage[linesnumbered, ruled,vlined, norelsize]{algorithm2e}
\usepackage{multirow}
\usepackage{booktabs,array}
\usepackage[compact]{titlesec}
\usepackage{pifont}
\newcommand{\cmark}{\ding{51}}
\newcommand{\xmark}{\ding{55}}

\newtheorem{thm}{\textbf{Theorem}}
\newtheorem{defn}{\textbf{Definition}}

\setlist{nolistsep,leftmargin=*}
\newcolumntype{P}[1]{>{\centering\arraybackslash}p{#1}}

\AtBeginDocument{%
  \providecommand\BibTeX{{%
    \normalfont B\kern-0.5em{\scshape i\kern-0.25em b}\kern-0.8em\TeX}}}

\setcopyright{iw3c2w3}
\copyrightyear{2020}
\acmYear{2020}
\acmDOI{10.1145/3366423.3380201}

\acmConference[WWW '20]{Proceedings of The Web Conference 2020}{April 20--24, 2020}{Taipei, Taiwan}
\acmBooktitle{Proceedings of The Web Conference 2020 (WWW '20), April 20--24, 2020, Taipei, Taiwan} 
\acmPrice{}
\acmISBN{978-1-4503-7023-3/20/04}

\begin{document}

\title{GraphGen: A Scalable Approach to Domain-agnostic Labeled Graph Generation}

\author{Nikhil Goyal}
\authornote{Both authors contributed equally to this research.}
\affiliation{
  \institution{Indian Institute of Technology, Delhi}
  \texttt{nikhil10sep@gmail.com}
}

\author{Harsh Vardhan Jain}
\authornotemark[1]
\affiliation{
  \institution{Indian Institute of Technology, Delhi}
  \texttt{harshvardhanjain7@gmail.com}
}

\author{Sayan Ranu}
\affiliation{
  \institution{Indian Institute of Technology, Delhi}
  \texttt{sayanranu@cse.iitd.ac.in}
}


\begin{abstract}
  Graph generative models have been extensively studied in the data mining literature. While traditional techniques are based on generating structures that adhere to a pre-decided distribution, recent techniques have shifted towards \emph{learning} this distribution directly from the data. While learning-based approaches have imparted significant improvement in quality, some limitations remain to be addressed. First, learning graph distributions introduces additional computational overhead, which limits their scalability to large graph databases. Second, many techniques only learn the structure and do not address the need to also learn node and edge labels, which encode important semantic information and influence the structure itself. Third, existing techniques often incorporate domain-specific rules and lack generalizability. Fourth, the experimentation of existing techniques is not comprehensive enough due to either using weak evaluation metrics or focusing primarily on synthetic or small datasets. In this work, we develop a \emph{domain-agnostic} technique called \textsc{GraphGen} to overcome all of these limitations. \textsc{GraphGen} converts graphs to sequences using \emph{minimum DFS codes}. Minimum DFS codes are \emph{canonical labels} and capture the graph structure precisely along with the label information. The complex joint distributions between structure and semantic labels are learned through a novel LSTM architecture. Extensive experiments on million-sized, real graph datasets show \textsc{GraphGen} to be $4$ times faster on average than state-of-the-art techniques while being significantly better in quality across a comprehensive set of $11$ different metrics. Our code is released at: \url{https://github.com/idea-iitd/graphgen}.
\end{abstract}




\maketitle

\section{Introduction and Related Work}
\label{sec:related_work}
Modeling and generating real-world graphs have applications in several domains, such as understanding interaction dynamics in social networks\cite{graphgan, graphite, deepnc}, graph classification\cite{graphsig_jcim, pgraphsig, mantra}, and anomaly detection\cite{graphsig}. Owing to their wide applications, development of generative models for graphs has a rich history, and many methods have been proposed. However, a majority of the techniques, make assumptions about certain properties of the graph database, and generate graphs that adhere to those assumed properties\cite{ermodel, bamodel, mmsb, kroneckergraphs, dvae}. A key area that lacks development is the ability to directly \emph{learn} patterns, both local as well as global, from a given set of observed graphs and use this knowledge to generate graphs instead of making prior assumptions. In this work, we bridge this gap by developing a domain-agnostic graph generative model for labeled graphs without making any prior assumption about the dataset. Such an approach reduces the usability barrier for non-technical users and applies to domains where distribution assumptions made by traditional techniques do not fit well, \textit{e.g.} chemical compounds, protein interaction networks \textit{etc}. 

Modeling graph structures is a challenging task due to the inherent complexity of encoding both local and global dependencies in link formation among nodes. We briefly summarize how existing techniques tackle this challenge and their limitations.

\begin{table}[t]
\centering
\caption{Limitations of existing graph generative models. $n$ and $m$ denote the number of nodes and edges respectively. Any other variable is a hyper-parameter of the respective model. We consider a model scalable if its complexity is linear in $n$ and $m$.}
\vspace{-0.1in}
\label{tab:limitations}
\scalebox{0.88} {
{\scriptsize
\begin{tabular}{|l|c|c|c|cc|}
\hline
{\bf Technique} & {\bf Domain-agnostic} & {\bf Node Labels} & {\bf Edge Labels} & {\bf Scalable} & {\bf Complexity}\\
\hline
MolGAN\cite{molgan} & \xmark & \cmark & \cmark & \xmark & $O(n^2)$\\
\hline
NeVAE\cite{nevae} & \xmark & \cmark & \cmark & \cmark  & $O(Sm)$\\
\hline
GCPN\cite{gcpn} & \xmark & \cmark & \cmark & \xmark  &O$(m^2)$\\
\hline
LGGAN\cite{lggan} & \cmark & \cmark & \xmark & \xmark & $O(n^2)$ \\
\hline
Graphite\cite{graphite} & \cmark & \cmark & \xmark & \xmark  &$O(n^2)$\\
\hline
DeepGMG\cite{deepgmg} & \cmark & \cmark & \cmark & \xmark  & $O(n^2m)$\\
\hline
GraphRNN\cite{graphrnn} & \cmark & \xmark & \xmark & \cmark  & $O(nM)$\\
\hline 
GraphVAE\cite{graphvae} & \cmark & \cmark & \cmark & \xmark & $O(n^4)$ \\
\hline
GRAN\cite{gran} & \cmark & \xmark & \xmark & \xmark & $O(n(m + n))$ \\

\hline
\hline
\textbf{\textsc{GraphGen}} & \cmark & \cmark & \cmark & \cmark  & $O(m)$ \\
\hline

\end{tabular}}
}
\end{table}

\textbf{Traditional graph generative models: } Several models exist\cite{mmsb, bamodel, ermodel, exprandomgraphs, kroneckergraphs, smallworldnetwork} that are engineered towards modeling a pre-selected family of graphs, such as random graphs\cite{ermodel}, small-world networks \cite{smallworldnetwork}, and scale-free graphs\cite{bamodel}. 
The apriori assumption introduces several limitations. First, due to pre-selecting the family of graphs, \textit{i.e.} distributions modeling some structural properties, they cannot adapt to datasets that either do not fit well or evolve to a different structural distribution. Second, these models only incorporate structural properties and do not look into the patterns encoded by labels. 
Third, these models assume the graph to be homogeneous. In reality, it is possible that different local communities (subgraphs) within a graph adhere to different structural distributions. 

\textbf{Learning-based graph generative models: } To address the above outlined limitations, recent techniques have shifted towards a learning-based approach\cite{deepgmg, graphrnn, lggan, graphvae, gram, gran}. 
While impressive progress has been made through this approach, there is scope to further improve the performance of graph generative models. We outline the key areas we target in this work.

\begin{itemize}
\item \textbf{Domain-agnostic modeling: } Several models have been proposed recently that target graphs from a specific domain and employ domain-specific constraints\cite{nevae, molrnn, molgan, gcpn, jtvae, constrainedgraphvae, multiobjectuvedrugdesign} in the generative task. 
 While these techniques produce excellent results on the target domain, they do not generalize well to graphs from other domains.

\item \textbf{Labeled graph generation: } 
Real-world graphs, such as protein interaction networks, chemical compounds, and knowledge graphs, encode semantic information through node and edge labels. To be useful in these domains, we need a generative model that jointly captures the relationship between structure and labels (both node and edge labels). 

\item \textbf{Data Scalability: }It is not uncommon to find datasets containing millions of graphs\cite{zinc}. Consequently, it is important for any generative model to scale to large training datasets so that all of the information can be exploited for realistic graph generation. Many of the existing techniques do not scale to large graph databases\cite{graphrnn, deepgmg, gran, lggan, graphvae}. Often this non-scalability stems from dealing with exponential representations of the same graph, complex neural architecture, and large parameter space. For example, LGGAN\cite{lggan} models graph through its adjacency matrix and therefore has $O(n^2)$ complexity, where $n$ is the number of nodes in the graph. 
GraphRNN\cite{graphrnn} is the most scalable among existing techniques due to employing a smaller recurrent neural network (RNN) of $O(nM)$ complexity, where $M$ is a hyper-parameter. In GraphRNN, the sequence representations are constructed by performing a large number of breadth-first-search (BFS) enumerations for each graph. Consequently, even if the size of the graph dataset is small, the number of unique sequences fed to the neural network is much larger, which in turn affects scalability.
\end{itemize}

\begin{figure}[t]
  \centering
  \includegraphics[width=3.0in]{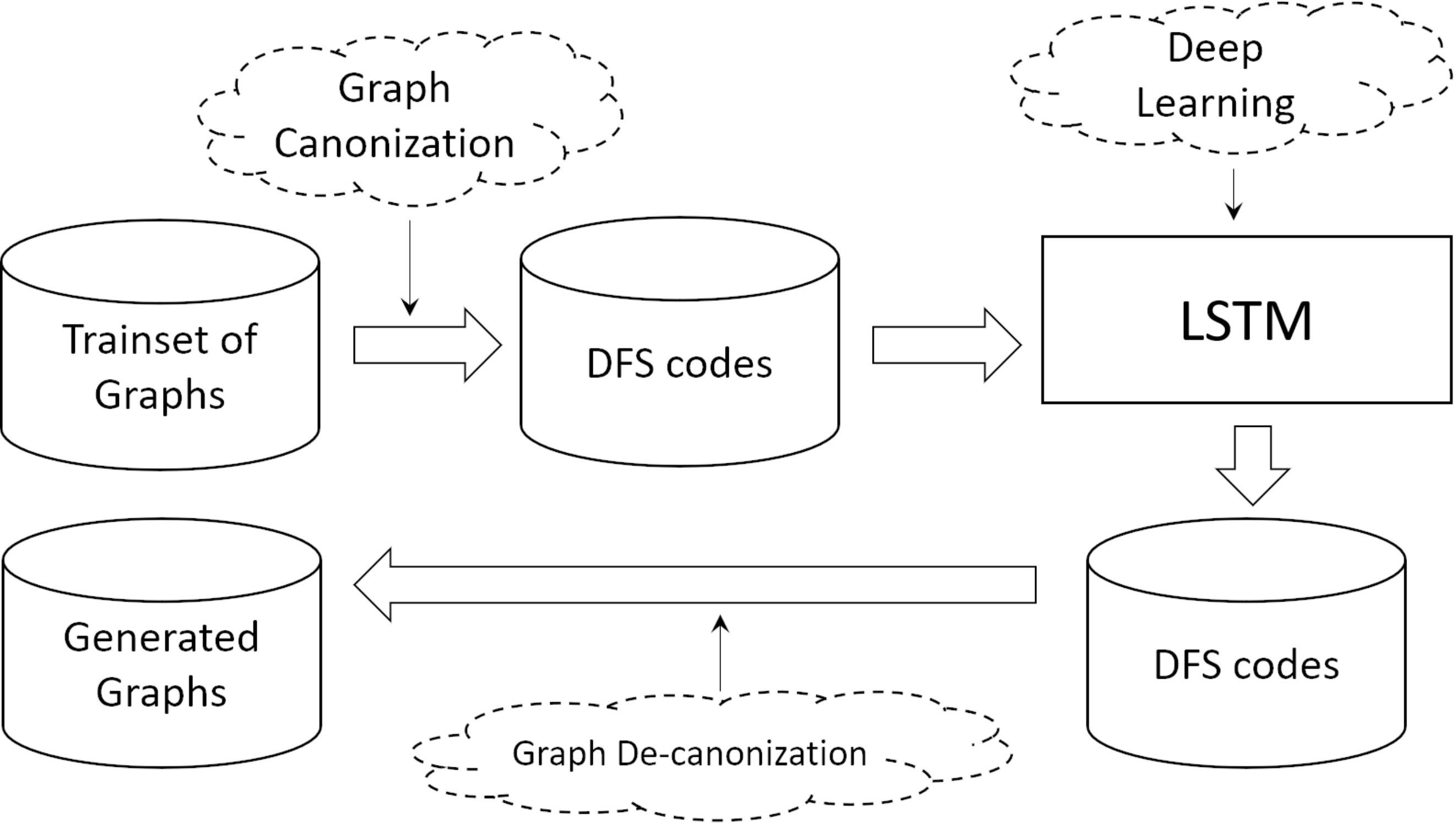}
  \caption{Flowchart of \textsc{GraphGen}.}
  \label{fig:flowchart}
\vspace{-0.10in}
\end{figure}

Table~\ref{tab:limitations} summarizes the limitations of learning-based approaches. \textsc{GraphGen} addresses all of the above outlined limitations and is the first technique that is domain-agnostic, assumption-free, models both labeled and unlabeled graphs, and data scalable. Specifically, we make the following contributions.
\begin{itemize}

\item We propose the problem of learning a graph generative model that is assumption-free, domain-agnostic, and captures the complex interplay between semantic labels and graph structure (\S~\ref{sec:formulation}). 

\item To solve the proposed problem, we develop an algorithm called \textsc{GraphGen} that employs the unique combination of \emph{graph canonization} with deep learning. Fig.~\ref{fig:flowchart} outlines the pipeline. Given a database of training graphs, we first construct the canonical label of each graph\cite{fsg, gspan}. The canonical label of a graph is a string representation such that two graphs are \emph{isomorphic} if and only if they have the same canonical label. We use \emph{DFS codes} for canonization. Through canonization, the graph generative modeling task converts to a sequence modeling task, for which we use Long Short-term Memory (LSTM)\cite{lstm} (\S~\ref{sec:graphgen}).


\item We perform an extensive empirical evaluation on real million-sized graph databases spanning three different domains. Our empirical evaluation establishes that we are significantly better in quality than the state-of-the-art techniques across an array of metrics, more robust, and $4$ times faster on average (\S~\ref{sec:experiments}).
\end{itemize}

\section{Problem Formulation}
\label{sec:formulation}

As a notational convention, a graph is represented by the tuple $G = (V, E)$ consisting of node set $V = \{v_{1},\cdots, v_{n}\}$ and edge set $E = \{(v_{i}, v_{j}) | v_{i}, v_{j} \in V \}$. Graphs may be annotated with node and edge labels. Let $\mathcal{L}_n:V \to \mathbb{V}$  and $\mathcal{L}_e: E \to \mathbb{E} $ be the node and edge label mappings respectively where $\mathbb{V}$ and $\mathbb{E}$ are the set of all node and edge labels respectively. We use the notation $L(v)$ and $L(e)$ to denote the labels of node $v$ and edge $e$ respectively. We assume that the graph is connected and there are no self-loops.

The goal of a graph generative model is to learn a distribution $p_{model}(\mathbb{G})$ over graphs, from a given set of observed graphs $\mathbb{G} = \{G_1, G_2,\cdots, G_m\}$ that is drawn from an underlying hidden distribution $p(\mathbb{G})$. Each $G_i$ could have a different number of nodes and edges, and could possibly have a different set of node labels and edge labels. The graph generative model is effective if the learned distribution is similar to the hidden distribution, \textit{i.e.}, $p_{model}(\mathbb{G})\approx p(\mathbb{G})$. More simply, the learned generative model should be capable of generating \emph{similar} graphs of varied sizes from the same distribution as $p(\mathbb{G})$ without any prior assumption about structure or labeling. 

\subsection{Challenges}

\begin{itemize}
\item \textbf{Large and variable output space:} The structure of a graph containing $n$ nodes can be characterized using an $n\times n$ adjacency matrix, which means a large output space of $O(n^2)$ values. With node and edge labels, this problem becomes even more complex as the adjacency matrix is no longer binary. Furthermore, the mapping from a graph to its adjacency matrix is not one-to-one. A graph with $n$ nodes can be equivalently represented using $n!$ adjacency matrices corresponding to each possible node ordering. Finally, $n$ itself varies from graph to graph. 

\item \textbf{Joint distribution space: }  While we want to learn the hidden graph distribution $p(G)$, defining $p(G)$ itself is a challenge since graph structures are complex and composed of various properties such as node degree, clustering coefficients, node and edge labels, number of nodes and edges, \textit{etc}. One could learn a distribution over each of these properties, but that is not enough since these distributions are not independent. The key challenge is therefore to learn the complex dependencies between various properties.

\item \textbf{Local and global dependencies: } The joint distribution space itself may not be homogeneous since the dependence between various graph parameters varies across different regions of a graph. For example, not all regions of a graph are equally dense. 

\end{itemize}

We overcome these challenges through \textsc{GraphGen}.

\begin{figure}
\centering
\subfigure[Graph]{
\label{fig:graph}
\includegraphics[width=0.6in]{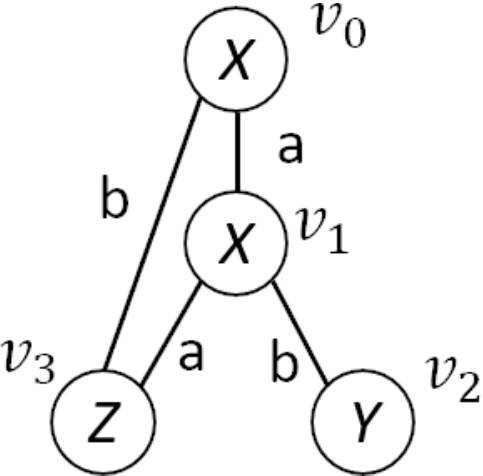}
}
\subfigure[DFS traversal on the graph]{
\label{fig:dfs}
\includegraphics[width=1.2in]{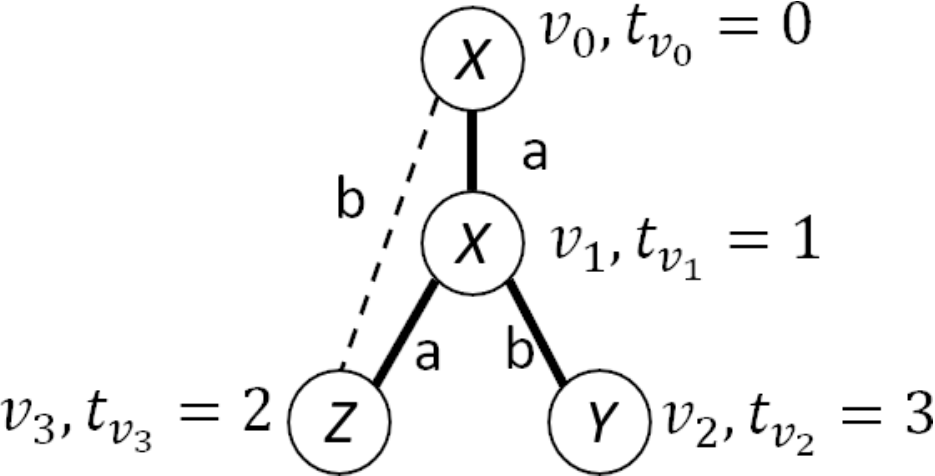}
}
\subfigure[DFS traversal on the graph]{
\label{fig:dfs1}
\includegraphics[width=1.2in]{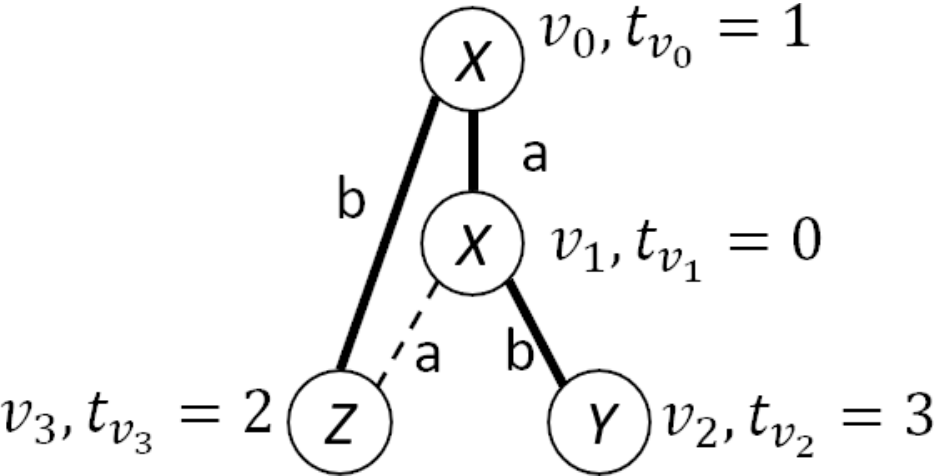}
}
\vspace{-0.1in}
\caption{Illustrates the DFS code generation process.}
\label{fig:dfsex}
\end{figure}
\section{\textsc{GraphGen}: Generative model for labeled graphs}
\label{sec:graphgen}

Instead of using adjacency matrices to capture graph structures, we perform graph canonization to convert each graph into a unique sequence. This step provides two key advantages. First, the length of the sequence is $|E|$ for graph $G(V,E)$ instead of $|V|^2$. Although in the worst case, $|E|=|V|^2$, in real graphs, this is rarely seen. Second, there is a one-to-one mapping between a graph and its canonical label instead of $|V|!$ mappings. Consequently, the size of the output space is significantly reduced, which results in faster and better modeling.

\subsection{DFS Codes: Graphs to Sequences} 
\label{sec:graphstoseqs}

First, we formally define the concept of \emph{graph canonization}.
\begin{defn}[Graph Isomorphism]
\label{def:isomorphism}
Graph $G_1=(V_1,$ $E_1)$ is \emph{isomorphic} to
$G_2=(V_2,E_2)$ if there exists a bijection $\phi$ such that
for every vertex $v \in V_1,\; \phi(v) \in V_2$ and for every edge $e =
(u, v) \in E_1, \phi(e)=(\phi(u), \phi(v)) \in  E_2$. If the graphs are labeled, we also need to ensure that the labels of mapped nodes and edges are the same, \textit{i.e.}, $\mathcal{L}(v)=\mathcal{L}(\phi(v))$ and $\mathcal{L}(e)=\mathcal{L}(\phi(e))$.

\end{defn}

\begin{defn}[Canonical label]
The canonical label of a graph is a string representation such that two graphs have the same label if and only if they are isomorphic to each other.
\end{defn}



We use \emph{DFS codes}\cite{gspan} to construct graph canonical labels. DFS code encodes a graph into a unique edge sequence by performing a depth first search (DFS). The DFS traversal may start from any node in the graph. To illustrate, consider Fig.~\ref{fig:graph}. A DFS traversal starting at $v_0$ on this graph is shown in Fig.~\ref{fig:dfs} where the bold edges show the edges that are part of the DFS traversal. During the DFS traversal, we assign each node a \emph{timestamp} based on when it is discovered; the starting node has the timestamp $0$. Fig.~\ref{fig:dfs} shows the timestamps if the order of DFS traversal is $(v_0,v_1),(v_1,v_3), (v_1,v_2)$. We use the notation $t_u$ to denote the timestamp of node $u$. Any edge $(u,v)$ that is part of the DFS traversal is guaranteed to have $t_u<t_v$, and we call such an edge a \emph{forward edge}. On the other hand, the edges that are not part of the DFS traversal, such as $(v_3,v_0)$ in Fig.~\ref{fig:dfs}, are called \emph{backward edges}. In Fig.~\ref{fig:dfs}, the bold edges represent the forward edges, and the dashed edge represents the backward edge. Based on the timestamps assigned, an edge $(u,v)$ is described using a $5$-tuple $(t_u,t_v,L_u,L_{(u,v)},L_v)$, where $L_u$ and $L_{(u,v)}$ denote the node and edge labels of $u$ and $(u,v)$ respectively. 

Given a DFS traversal, our goal is to impose a \emph{total ordering} on all edges of the graph. Forward edges already have an ordering between them based on when they are traversed in DFS. To incorporate backward edges, we enforce the following rule:
\begin{itemize}
\item Any backward edge $(u,u')$ must be ordered before all forward edges of the form $(u,v)$.
\item Any backward edge $(u,u')$ must be ordered after the forward edge of the form $(w,u)$, \textit{i.e.}, the first forward edge pointing to $u$.
\item Among the backward edges from the same node $u$ of the form $(u,u')$ and $(u,u'')$, $(u,u'')$ has a higher order if $t_{u''}<t_{u'}$.
\end{itemize}

With the above rules, corresponding to any DFS traversal of a graph, we can impose a total ordering on all edges of a graph and can thus convert it to a sequence of edges that are described using their $5$-tuple representations. We call this sequence representation a \emph{DFS code}.

\begin{example}
Revisiting Fig.~\ref{fig:dfs}, the DFS code is $\langle (0,1,X,a,X)$ $(1,2,X,a,Z)\:(2,0,Z,b,X)\:(1,3,X,b,Y)\rangle$. A second possible DFS traversal is shown in Fig.~\ref{fig:dfs1}, for which the edge sequence is $\langle (0,1,X,a,X)$ $(1,2,X,b,Z)\:(2,0,Z,a,X)\:(0,3,X,b,Y)\rangle$.
\end{example}

Since each graph may have multiple DFS traversals, we choose the \emph{lexicographically smallest} DFS code based upon the lexicographical ordering proposed in\cite{gspan}. Hereon, the lexicographically smallest DFS code is referred to as the \emph{minimum DFS code}. We use the notation $\mathcal{F}(G)= S = [s_1,\cdots,s_m]$ to denote the \emph{minimum DFS code} of graph $G$, where $m = |E|$.

\begin{thm}
$\mathcal{F}(G)$ is a canonical label for any graph $G$.\cite{gspan}
\end{thm}

\begin{example}
The DFS code for Fig.~\ref{fig:dfs} is smaller than Fig.~\ref{fig:dfs1} since the second edge tuple $(1,2,X,a,Z)$ of Fig.~\ref{fig:dfs} is smaller than $(1,2,X,b,Z)$ of Fig.~\ref{fig:dfs1} as $a<b$. Since the first edge tuple in both sequences are identical the tie is broken through the second tuple.
\end{example}

\subsubsection{Computation Cost and Properties}
\label{sec:dfscost}
Constructing the minimum DFS code of a graph is equivalent to solving the graph isomorphism problem. Both operations have a worst-case computation cost of $O(|V|!)$ since we may need to evaluate all possible permutations of the nodes to identify the lexicographically smallest one. Therefore an important question arises: \textit{How can we claim scalability with a factorial computation complexity? } To answer this question, we make the following observations.
\begin{itemize}
\item \textbf{Labeled graphs are ubiquitous: } Most real graphs are labeled. Labels allow us to drastically prune the search space and identify the minimum DFS code quickly. For example, in Fig.~\ref{fig:graph}, the only possible starting nodes are $v_0$ or $v_1$ as they contain the lexicographically smallest node label ``X''. Among them, their 2-hop neighborhoods are enough to derive that $v_0$ must be the starting node of the minimum DFS code as $(1,2,X,a,Z)$ is the lexicographically smallest possible tuple after $(0,1,X,a,X)$ in the graph in Fig.~\ref{fig:graph}. We empirically substantiate this claim in \S~\ref{sec:impact_graphsize}.

\item \textbf{Invariants: } What happens if the graph is unlabeled or has less diversity in labels? In such cases, \emph{vertex and edge invariants} can be used as node and edge labels. Invariants are properties of nodes and edges that depend only on the structure and not on graph representations such as adjacency matrix or edge sequence. Examples include node degree, betweenness centrality, clustering coefficient, etc. We empirically study this aspect further in \S~\ref{sec:invariants}.

\item \textbf{Precise training:} Existing techniques rely on the neural network to learn the multiple representations that correspond to the same graph. 
Many-to-one mappings introduce impreciseness and bloat the modeling task with redundant information. In \textsc{GraphGen}, we feed precise one-to-one graph representations, which allows us to use a more lightweight neural architecture. Dealing with multiple graph representations are handled algorithmically, which leads to significant improvement in quality and scalability.
\end{itemize}

\subsection{Modeling Sequences}
\subsubsection{Model Overview}
Owing to the conversion of graphs to minimum DFS codes, and given that DFS codes are canonical labels, modeling a database of graphs $\mathbb{G}=\{G_1,\cdots, G_n\}$ is equivalent to modeling their sequence representations $\mathbb{S} = \{\mathcal{F}(G_1), \cdots, \mathcal{F}(G_n)\} = \{S_1,\cdots,S_n\}$, \textit{i.e.}, $p_{model}(\mathbb{G}) \equiv p_{model}(\mathbb{S})$. To model the sequential nature of the DFS codes, we use an \emph{auto-regressive} model to learn $p_{model}(\mathbb{S})$. At inference time, we sample DFS codes from this distribution instead of directly sampling graphs. Since the mapping of a graph to its minimum DFS code is a \emph{bijection}, \textit{i.e.}, $G=\mathcal{F}^{-1}(\mathcal{F}(G))$, the graph structure along with all node and edge labels can be fully constructed from the sampled DFS code. 
 
\begin{figure}[t]
\centering
\includegraphics[width=\linewidth]{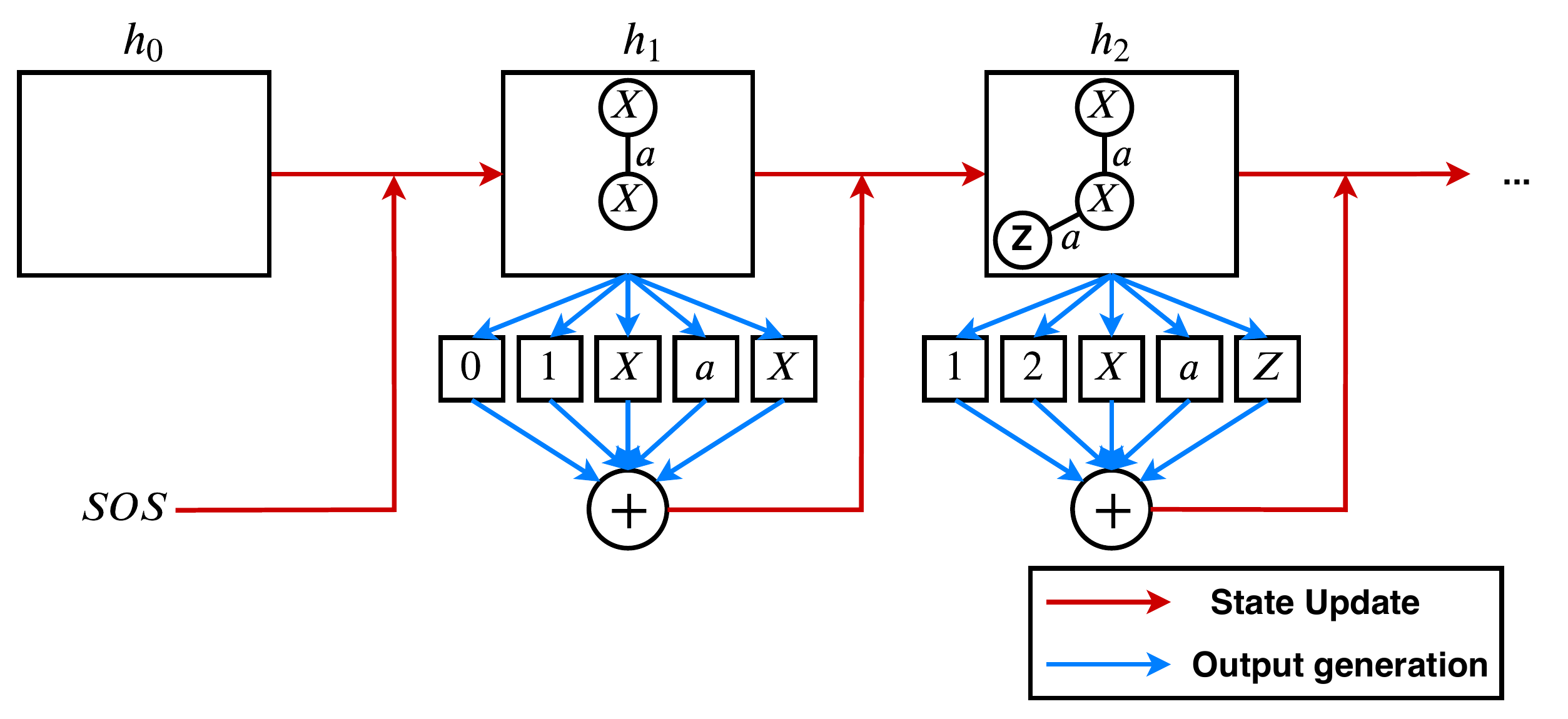}
\vspace{-0.1in}
\caption{Architecture of \textsc{GraphGen}. Red arrows indicate data flow in the RNN whose hidden state $h_i$ captures the state of the graph generated so far. Blue arrows show the information flow to generate the new edge tuple $s_i$.}
 
  \label{fig:architecture}
\end{figure}

Fig.\ref{fig:architecture} provides an overview of \textsc{GraphGen} model. For graph $G(V,E)$ with $\mathcal{F}(G) = S = [s_1,\cdots,s_m]$, the proposed auto-regressive model generates $s_i$ sequentially from $i = 1$ to $i = m = |E|$. This means that our model produces a single edge at a time, and since it belongs to a DFS sequence, it can only be between two previously generated nodes (backward edge) or one of the already generated nodes and an unseen node (forward edge), in which case a new node is also formed. Consequently, only the first generated edge produces two new nodes, and all subsequent edges produce at most one new node.

\subsubsection{Model Details}
We now describe the proposed algorithm to characterize $p(S)$ in detail. Since $S$ is sequential in nature, we decompose the probability of sampling $S$ from $p(\mathbb{S})$ as a product of \emph{conditional distribution} over its elements as follows:
\begin{equation}
    \label{eq:conditionaldecomp}
    p(S) = \prod_{i = 1}^{m + 1}p(s_i|s_1,\cdots,s_{i-1}) =  \prod_{i = 1}^{m + 1}p(s_i|s_{<i})
\end{equation}

where $m$ is the number of edges and $s_{m+1}$ is \emph{end-of-sequence} token EOS to allow variable length sequences. We denote $p(s_i|s_1,\cdots, \allowbreak s_{i-1})$ as $p(s_i| s_{<i})$ in further discussions. 

Recall, each element $s_i$ is an edge tuple of the form $(t_u,t_v,L_{u},L_{e}, \allowbreak L_{v})$, where $e=(u,v)$. We make the simplifying assumption that timestamps $t_u$, $t_v$, node labels $L_{u}$, $L_{v}$ and edge label $L_{e}$ at each generation step of $s_i$ are independent of each other. This makes the model tractable and easier to train. Note that this assumption is not dependent on data and hence is not forming prior bias of any sort on data. Mathematically, Eq.~\ref{eq:conditionaldecomp} reduces to the follows.

\begin{equation}\label{eq:expandeddist}
\begin{split}
    p(s_i|s_{<i}) &= p((t_u,t_v,L_u,L_{e},L_v))\:|\:s_{<i}) \\
    &= p(t_u|s_{<i}) \: \times \: p(t_v|s_{<i}) \: \times \: p(L_u|s_{<i}) \\
    & \qquad \times \: p(L_{e}|s_{<i}) \: \times \: p(L_v|s_{<i})
\end{split}
\end{equation}

Eq.~\ref{eq:expandeddist} is extremely complex as it has to capture complete information about an upcoming edge, \textit{i.e.}, the nodes to which the edge is connected, their labels, and the label of the edge itself. To capture this highly complex nature of $p(s_i|s_{<i})$ we propose to use expressive neural networks. \emph{Recurrent neural networks (RNNs)} are capable  of learning features and long term dependencies from sequential and time-series data\cite{rnnsurvey}. Specifically, LSTMs\cite{lstm} have been one of the most popular and efficient methods for reducing the effects of vanishing and exploding gradients in training RNNs. In recent times, LSTMs and \emph{stacked} LSTMs have delivered promising results in cursive handwriting\cite{handwrittingrnn}, sentence modeling\cite{sentencevae} and, speech recognition\cite{stackedlstm}. 

In this paper, we design a custom LSTM that consists of a \emph{state transition function} $f_{trans}$ (Eq.~\ref{eq:trans}), an embedding function $f_{emb}$ (Eq.~\ref{eq:trans}), and five separate \emph{output functions} (Eqs.~\ref{eq:tu}-\ref{eq:lv}) for each of the five components of the $5$-tuple $s_i=(t_u,t_v,L_u,L_{e},L_v)$.
($\sim_M$ in the following equations represents sampling from multinomial distribution)
\begin{align}
\label{eq:trans}
    h_i & = f_{trans}(h_{i-1},f_{emb}(s_{i-1})) \\
\label{eq:tu}
    t_u & \sim_M \; \theta_{t_u} = f_{t_u}(h_i) \\
\label{eq:tv}
    t_v & \sim_M \; \theta_{t_v} = f_{t_v}(h_i) \\
\label{eq:lu}
    L_u & \sim_M \; \theta_{L_u} = f_{L_u}(h_i) \\
\label{eq:le}
    L_e & \sim_M \; \theta_{L_e} = f_{L_e}(h_i) \\
\label{eq:lv}
    L_v & \sim_M \; \theta_{L_u} = f_{L_v}(h_i) \\
\label{eq:si}
    s_i & = \mathsf{concat}(t_u,t_v,L_u,L_e,L_v)
\end{align}

Here, $s_i \in \{0,1\}^k$ is the concatenated component wise \emph{one-hot encoding} of the real edge, and $h_i \in \mathbb{R}^d$ is an LSTM hidden state vector that encodes the state of the graph generated so far. An embedding function $f_{emb}$ is used to compress the sparse information possessed by the large edge representation ($s_i$) to a small vector of real numbers (Eq.~\ref{eq:trans}). Given the graph state, $h_i$, the output of the functions $f_{t_u}(h_i)$, $f_{t_v}(h_i)$, $f_{L_u}(h_i)$, $f_{e}(h_i)$, $f_{L_v}(h_i)$ \textit{i.e.} $\theta_{t_u}$, $\theta_{t_v}$, $\theta_{L_u}$, $\theta_{e}$, $\theta_{L_v}$ respectively represent the multinomial distribution over possibilities of five components of the newly formed edge tuple. Finally, components of the newly formed edge tuple are sampled from the multinomial distributions parameterized by respective $\theta$s and concatenated to form the new edge ($s_i$) (Eq.~\ref{eq:si}). 

With the above design, the key components that dictate the modeling quality are the multinomial probability distribution parameters $\theta_{t_u}$, $\theta_{t_v}$, $\theta_{L_u}$, $\theta_{e}$, $\theta_{L_v}$ over each edge tuple $t_u$, $t_v$, $L_u$, $L_e$, $L_v$. Our goal is, therefore, to ensure that the learned distributions are as close as possible to the real distributions in the training data. Using the broad mathematical architecture explained above, we will now explain the Graph Generation process followed by the sequence modeling / training process.

\begin{algorithm}[b]
\DontPrintSemicolon
\scriptsize
\SetAlgoLined
\SetKwInOut{Input}{Input}
\SetKwInOut{Output}{Output}
\Input{LSTM based state transition function $f_{trans}$, embedding function $f_{emb}$, five output functions $f_{t_u},f_{t_v},f_{L_u},f_{L_e},f_{L_v}$, empty graph state $h_0$, start token $SOS$, and end token $EOS$}
\Output{Graph $G$}
 
$S \leftarrow \emptyset$\; \label{inf:init:S}
$\widehat{s}_0 \leftarrow SOS$\; \label{inf:init:s}
$i \leftarrow 0$\; \label{inf:init:i}
\Repeat{$\exists \: c_i, c_i=EOS$ }{
  $i \leftarrow i + 1$\;
  $h_i \leftarrow f_{trans}(h_{i-1}, f_{emb}(\widehat{s}_{i-1}))$\;
  $\widehat{s_i}\leftarrow \emptyset$\;
  \For {\label{inf:loop:s}$c \in \{t_u,t_v,L_u,L_e,L_v\}$}{
    $ \theta_{c} \leftarrow f_c(h_i) $\;
    \tcp{sample $c_i$ from Multinomial distribution parameterized by $\theta_{c}$}
    $c_i \sim_{M} {\theta_{c}}$\;
    $\widehat{s_i}\leftarrow \mathsf{concat}(\widehat{s_i},c_i)$\;
  }
  $S\leftarrow S || \langle \widehat{s_i} \rangle$ \label{inf:update:S}\tcp*[f]{Appending $\widehat{s_i}$ to sequence $S$}
}
\Return {$\mathcal{F}^{-1}(S)$} \tcp*[f]{$\mathcal{F}^{-1}$ is mapping from minimum DFS code to graph}
\caption{Graph generation algorithm}
\label{alg:inference}
\end{algorithm}

\subsubsection{Graph Generation} Given the learned state transition function $f_{trans}$, embedding function $f_{emb}$ and output functions $f_{t_u}$, $f_{t_v}$, $f_{L_u}$, $f_{L_e}$, $f_{L_v}$, Alg.~\ref{alg:inference} provides the pseudocode for the graph generation process. In Alg.~\ref{alg:inference}, $S$ stores the generated sequence and is initialized along with other variables (line \ref{inf:init:S}-\ref{inf:init:i}). $\widehat{s_i} \in \{0,1\}^k$, which represents $i^{th}$ edge tuple formed for sequence $S$, is component-wise sampled from the multinomial distributions learned by output functions (loop at line \ref{inf:loop:s}) and finally appended to the sequence $S$ (line \ref{inf:update:S}). This process of forming a new edge and appending it to sequence $S$ is repeated until any of the five components is sampled as an EOS. Note, since each of the five components of $\widehat{s_i}$ \textit{i.e.}, $c_i, c \in \{t_u,t_v,L_u,L_e,L_v\}$ are one-hot encodings and can be of different size, their EOS could be of different sizes, in which case they will be compared with their respective EOS to indicate the stopping condition. 
 
\subsubsection{Graph Modeling} We now discuss how the required inference functions (Eqs.~\ref{eq:trans}-\ref{eq:si}) are learned from a set of training graphs.

Given an input graph dataset $\mathbb{G}$, Alg.~\ref{alg:training} first converts the graph dataset $\mathbb{G}$ to a sequence dataset $\mathbb{S}$ of minimum DFS codes (line \ref{model:init:S}). We initialize the unlearned neural functions by suitable weights (line \ref{model:init:f}). Further details about these functions are given in Sec.~\ref{sec:parameters}. For each edge $s_i$ in each sequence $S \in \mathbb{S}$, a concatenated component-wise one-hot encoding of the real edge $s_i$ is fed to the embedding function $f_{emb}$, which compresses it and its compressed form along with previous graph state is passed to the transition function $f_{trans}$. $f_{trans}$ generates the new graph state $h_i$ (line \ref{model:update:h_i}). This new state $h_i$ is passed to all the five output functions that provide us the corresponding probability distributions over the components of the new edge tuple (line \ref{model:update:theta_c}). We concatenate these probability distributions to form $\widetilde{s_i} \in \mathbb{R}^k$ (same size as that of $s_i$), which is representative of a new edge (line \ref{model:update:s_i}). Finally, this component-wise concatenated probability distribution $\widetilde{s_i}$ is matched (component-wise) against the real tuple from the DFS code being learned. This implies that we use ground truth rather than the model's own prediction during successive time steps in training. In contrast to this, \textsc{GraphGen} uses its own predictions during graph generation time. The accuracy of the model is optimized using the cumulative \emph{binary cross-entropy} loss function. Here $s[c]$ represents component $c \in \{t_u,t_v,L_u,L_e,L_v\}$, and $log$ is taken elementwise on a vector. 

\begin{equation}
\label{eq:bce}
BCE(\widetilde{s_i}, {s_i})= -\sum_{c}\left( s_i[c]^T\log \widetilde{s}_i[c] + (1-s_i[c])^T\log (1-\widetilde{s}_i[c])\right)
\end{equation}

\begin{table}[t]
\caption{Variables and their sizes}
\vspace{-0.1in}
\label{tab:Vars}
{\footnotesize
\begin{tabular}{|c|c|}
    \hline
    Variable (One-hot vector)  & Dimension \\
    \hline
    $t_u, t_v$  &  $\max\limits_{\forall G(V,E)\in\mathbb{G}}{}  |V| + 1$ \\
    $L_u, L_v$ &  $ |\mathbb{V}| + 1$ \\
    $L_e$ &  $|\mathbb{E}| + 1$ \\
    \hline  
\end{tabular}}
\end{table}

\begin{algorithm}[b]
\DontPrintSemicolon
\scriptsize
\SetAlgoLined
\SetKwInOut{Input}{Input}
\SetKwInOut{Output}{Output}
\Input{Dataset of Graphs $\mathbb{G} = \{G_1,\cdots,G_n\}$}
\Output{Learned functions $f_{trans}, \:f_{t_u}, \:f_{t_v},\:f_{L_u}, \:f_{L_e}, \: f_{L_v}$, and embedding function $f_{emb}$}

$\mathbb{S}=\{S=\mathcal{F}(G)\:|\: \forall G\in\mathbb{G}\}$ \; \label{model:init:S}
Initialize $f_{t_u}, f_{t_v}, f_{L_u}, f_{L_e}, f_{L_v}, f_{emb}$\; \label{model:init:f}
\Repeat(\tcp*[f]{1 Epoch}){stopping criteria}{
  \For {$\forall S=[s_1,\cdots, s_m] \in \mathbb{S}$}{
    $s_0 \leftarrow SOS$ ; Initialize $h_0$ \;
    $loss \leftarrow 0$\;
    \For(\tcp*[f]{$s_{m+1}$ for EOS tokens}){ $i$ from $1$ to $m + 1$}{
      $h_i \leftarrow f_{trans}(h_{i-1}, f_{emb}(s_{i-1}))$\; \label{model:update:h_i}
      \tcp{$\widetilde{s_i}$ will contain component-wise probability distribution vectors of $\widehat{s_i}$}
      $\widetilde{s_i} \leftarrow \phi$\;
      \For {$c \in \{t_u,t_v,L_u,L_e,L_v\}$}{
        $ \theta_{c} \leftarrow f_c(h_i) $\; \label{model:update:theta_c}
        $\widetilde{s_i} \leftarrow \mathsf{concat}( \widetilde{s_i}, \theta_c)$\; \label{model:update:s_i}
      }
      \tcp{Note: $s_i$ consists of 5 components (See Eq.~\ref{eq:bce})}
      \emph{loss} $\leftarrow$ \emph{loss} + $BCE(\widetilde{s_i}, {s_i})$\;
    }
    Back-propagate loss and update weights\;
  }
}(\tcp*[f]{Typically when validation loss is minimized})
\caption{Graph modeling algorithm}
\label{alg:training}
\end{algorithm}

\subsubsection{Parameters} 
\label{sec:parameters}
State transition function, $f_{trans}$ in our architecture is a stacked LSTM-Cell, each of the five output functions $f_{t_u}$, $f_{t_v}$, $f_{L_u}$, $f_{L_e}$, $f_{L_v}$ are fully connected, independent \emph{Multi-layer Perceptrons (MLP)} with dropouts and $f_{emb}$ is a simple linear embedding function. The dimension of the one-hot encodings of $t_u$, $t_v$, $L_u$, $L_e$, $L_v$ are estimated from the training set. Specifically, the largest values for each element are computed in the first pass while computing the minimum DFS codes of the training graphs. Table \ref{tab:Vars} shows the dimensions of the one-hot encodings for each component of $s_i$. Note that each variable has one size extra than their intuitive size to incorporate the EOS token. 
Since we set the dimensions of $t_u$ and $t_v$ to the largest training graph size, in terms of the number of nodes, the largest generated graph cannot exceed the largest training graph. Similarly, setting the dimensions of $L_u$, $L_v$, and $L_e$ to the number of unique node and edge labels in the training set means that our model is bound to produce only labels that have been observed at least once. Since the sizes of each component of $s_i$ is fixed, the size of $s_i$, which is formed by concatenating these components, is also fixed, and therefore, satisfies the pre-requisite condition for LSTMs to work, \textit{i.e.}, each input must be of the same size. We employ weight sharing among all time steps in LSTM to make the model scalable.

\subsubsection{Complexity Analysis of Modeling and Inference.}
\label{sec:complexity}

Consistent with the complexity analysis of existing models (Recall Table~\ref{tab:limitations}), we only analyze the complexity of the forward and backward propagations. The optimization step to learn weights is not included. 

The length of the sequence representation for graph $G(V,E)$ is $|E|$. All operations in the LSTM model consume $O(1)$ time per edge tuple and hence the complexity is $O(|E|)$. The generation algorithm consumes $O(1)$ time per edge tuple and for a generated graph with edge set $E$, the complexity is $O(|E|)$.

\section{Experiments}
\label{sec:experiments}
In this section, we benchmark \textsc{GraphGen} on real graph datasets and establish that:
\begin{itemize}

\item \textbf{Quality: }\textsc{GraphGen}, on average, is significantly better than the state-of-the-art techniques of GraphRNN\cite{graphrnn} and DeepGMG\cite{deepgmg}.

\item \textbf{Scalability: }\textsc{GraphGen}, on average, is $4$ times faster than Graph\-RNN, which is currently the fastest generative model that works for labeled graphs. In addition, \textsc{GraphGen} scales better with graph size, both in quality and efficiency.
\end{itemize}

The code and datasets used for our empirical evaluation can be downloaded from \url{https://github.com/idea-iitd/graphgen}.

\begin{table}[t]
\caption{Summary of the datasets. \xmark \hspace{0.05cm} means labels are absent.}
\vspace{-0.1in}
\label{tab:datasets}
\scalebox{0.90} {
{\scriptsize
  \begin{tabular}{lllccccc}
    \toprule
    \# & Name & Domain & No. of graphs & $|V|$ & $|E|$ & $|\mathbb{V}|$ & $|\mathbb{E}|$\\
    \midrule
    1& NCI-H23 (Lung)\cite{cancer}& Chemical & 24k & [6, 50] & [6, 57] & 11 & 3 \\
    2& Yeast\cite{cancer}& Chemical & 47k & [5, 50] & [5, 57] & 11 & 3 \\
    3& MOLT-4 (Leukemia)\cite{cancer} & Chemical & 33k & [6, 111] & [6, 116] & 11 & 3 \\
    4&MCF-7 (Breast)\cite{cancer}& Chemical & 23k & [6, 111] & [6, 116] & 11 & 3 \\
    5&ZINC\cite{zinc} & Chemical & 3.2M & [3, 42] & [2, 49] & 9 & 4 \\
    6&Enzymes\cite{bogwartenzymes}& Protein & 575 & [2, 125] & [2, 149] & 3 & \xmark \\
    7&Citeseer\cite{sencitation} & Citation & 1 & 3312 & 4460 & 6 & \xmark \\
    8&Cora\cite{sencitation}& Citation & 1 & 2708 & 5278 & 6 & \xmark \\
  \bottomrule
\end{tabular}}
}
\end{table}

\begin{table*}
\caption{Summary of the performance by \textsc{GraphGen}, GraphRNN, and DeepGMG across $11$ different metrics measuring various aspects of quality, robustness and efficiency on $6$ datasets. The best performance achieved under each metric for a particular dataset is highlighted in bold font. Any value less than $10^{-3}$ is reported as $\approx 0$. The quality achieved by DeepGMG in Enzymes, Cora, and Citeseer are not reported since loss function for DeepGMG goes to $NAN$ and hence fails to scale.}
\vspace{-0.1in}
\label{tab:quality}
\scalebox{0.95}{
{\scriptsize
\begin{tabular}{|l|l|ccc|c|cc|ccc|cc|cc|}
  
    \hline
    Dataset & Model & Degree & Clustering & Orbit & NSPDK  & \makecell{Avg \# Nodes \\ (Gen/Gold)} & \makecell{Avg \# Edges \\ (Gen/Gold)} & \makecell{Node \\ Label} & \makecell{Edge \\ Label} & \makecell{Joint Node \\ Label \& Degree} & Novelty & Uniqueness & \makecell{Training \\ Time} & \# Epochs \\
    
    \hline
    \multirow{3}{4em}{Lung} & \textsc{GraphGen} & \textbf{0.009} &  $\mathbf{\approx 0}$ & $\mathbf{\approx 0}$ & \textbf{0.035} & \textbf{35.20/35.88} & \textbf{36.66/37.65}& \textbf{0.001} & $\mathbf{\approx 0}$ & \textbf{0.205} & $\mathbf{~\approx 100\%}$ & $\mathbf{\approx 100\%}$ & \textbf{6h} & 550 \\

    & GraphRNN & 0.103 & 0.301 & 0.043 & 0.325 & 6.32/35.88 & 6.35/37.65 & 0.193 & 0.005 & 0.836 & $86\%$ & $45\%$ & 1d 1h & 1900 \\

    & DeepGMG & 0.123 & 0.001 & 0.026 & 0.260 & 11.04/35.88 & 10.28/37.65 & 0.083 & 0.002 & 0.842 & $98\%$ & $98\%$ & 23h & \textbf{20} \\

    \hline
    \multirow{3}{4em}{Yeast} &\textsc{GraphGen} &\textbf{0.006} & $\mathbf{\approx 0}$ & $\mathbf{\approx 0}$ &\textbf{0.026} &\textbf{35.58/32.11} &\textbf{36.78/33.22} & \textbf{0.001} & $\mathbf{\approx 0}$ & \textbf{0.093} & \textbf{97\%} & \textbf{99\%} & \textbf{6h} & 250 \\

    & GraphRNN & 0.512 & 0.153 & 0.026 & 0.597 & 26.58/32.11 & 27.01/33.22 & 0.310 & 0.002 & 0.997 & 93\% & 90\% & 21h & 630 \\

    & DeepGMG & 0.056 & 0.002 & 0.008 & 0.239 & 34.91/32.11  & 35.08/33.22 &0.115  & $\mathbf{\approx 0}$ & 0.967 & 90\% & 89\% & 2d 3h & \textbf{18} \\
    
    \hline
    \makecell{\multirow{3}{1.5cm}{Mixed: Lung + Leukemia + Yeast + Breast}} & \textsc{GraphGen} &\textbf{0.005} & $\mathbf{\approx 0}$ & $\mathbf{\approx 0}$ &\textbf{0.023} &\textbf{37.87/37.6} &\textbf{39.24/39.14} & \textbf{0.001 } & $\mathbf{\approx 0}$ & \textbf{0.140} & \textbf{97\%} & \textbf{99\%} & \textbf{11h} & 350 \\

    & GraphRNN & 0.241 & $\mathbf{\approx 0}$ & 0.039 &0.331 & 8.19/37.6 & 7.2/39.14 & 0.102 & 0.010 & 0.879 & 62\% & 52\% & 23h & 400 \\

    & Deep GMG & 0.074 & 0.002 & 0.002 & 0.221 & 24.35/37.6 & 24.11/39.14 & 0.092 & 0.002 & 0.912 & 83\% & 83\% & 1d 23h & \textbf{11} \\
    
    \hline
    \multirow{2}{4em}{Enzymes} &\textsc{GraphGen} & 0.243 &0.198 &\textbf{0.016} & \textbf{0.051} & \textbf{32.44  /32.95} & \textbf{52.83/64.15} & \textbf{0.005}  & - & \textbf{0.249} & 98\% & \textbf{99\%} & \textbf{3h} & \textbf{4000} \\

    & GraphRNN & \textbf{0.090} & \textbf{0.151} & 0.038 &0.067 & 11.87/32.95 & 23.52/64.15 & 0.048 & - & 0.312 & \textbf{99\%} & 97\% & 15h & 20900 \\

    \hline
    \multirow{2}{4em}{Citeseer} &\textsc{GraphGen} &\textbf{0.089} & \textbf{0.083} & \textbf{0.100} & \textbf{0.020} & \textbf{35.99/48.56} & \textbf{41.81/59.14} & \textbf{0.024} & - & \textbf{0.032} & \textbf{83\%} & 95\% & \textbf{4h} & \textbf{400} \\

    & GraphRNN & 1.321 & 0.665 & 1.006 & 0.052 & 31.42/48.56 & 77.13/59.14 & 0.063 & - & 0.035 & 62\% & \textbf{100\%} & 12h & 1450 \\
 
    \hline
    \multirow{2}{4em}{Cora} & \textsc{GraphGen} & \textbf{0.061} & \textbf{0.117} & \textbf{0.089} & \textbf{0.012} & 48.66/58.39 & \textbf{55.82/68.61} & \textbf{0.017} & - & \textbf{0.013} & \textbf{91\%} & \textbf{98\%} & \textbf{4h} & \textbf{400} \\

    & GraphRNN & 1.125 & 1.002 & 0.427 & 0.093 & \textbf{54.01/58.39} & 226.46/68.61 & 0.085 & - & 0.015 & 70\% & 93\% & 9h & \textbf{400} \\

    \hline
\end{tabular}}
}
\end{table*}
\subsection {Experimental setup}
All experiments are performed on a machine running dual Intel Xeon Gold 6142 processor with 16 physical cores each, having 1 Nvidia 1080 Ti GPU card with 12GB GPU memory, and 384 GB RAM with Ubuntu 16.04 operating system. 

\subsubsection{Datasets}
\label{sec:dataset}
Table~\ref{tab:datasets} lists the various datasets used for our empirical evaluation. 
The semantics of the datasets are as follows.
\begin{itemize}
    \item \textbf{Chemical compounds:} The first five datasets are chemical compounds. We convert them to labeled graphs where nodes represent atoms, edges represent bonds, node labels denote the atom-type, and edge labels encode the bond order such as single bond, double bond, \textit{etc.}


    \item \textbf{Citation graphs:} Cora and Citeseer are citation networks; nodes correspond to publications and an edge represents one paper citing the other. Node labels represent the publication area. 

    \item \textbf{Enzymes:} This dataset contains protein tertiary structures representing $600$ enzymes from the BRENDA enzyme database\cite{brendaenzymes}. Nodes in a graph (protein) represent secondary structure elements, and two nodes are connected if the corresponding elements are interacting. The node labels indicate the type of secondary structure, which is either helices, turns, or sheets. This is an interesting dataset since it contains only three label-types. We utilize this dataset to showcase the impact of graph invariants. Specifically, we supplement the node labels with node degrees. For example, if a node has label ``A'' and degree $5$, we alter the label to ``5, A''.
\end{itemize}

\subsubsection{Baselines}
We compare \textsc{GraphGen} with DeepGMG\cite{deepgmg}\footnote{We modified the DeepGMG implementation for unlabeled graphs provided by \href{https://docs.dgl.ai/tutorials/models/3_generative_model/5_dgmg.html}{Deep Graph Library (DGL)}} and GraphRNN\cite{graphrnn}. For GraphRNN, we use the original code released by the authors. Since this code does not support labels, we extend the model to incorporate node and edge labels based on the discussion provided in the paper\cite{graphrnn}.
\subsubsection{Parameters and Training sets}
 For both GraphRNN and DeepGMG, we use the parameters recommended in the respective papers. Estimation of $M$ in GraphRNN and evaluation of DFS codes\footnote{We adapted Minimum DFS code implementation from \href{https://github.com/kaviniitm/DFSCode}{kaviniitm}} from graph database in \textsc{GraphGen} is done in parallel using 48 threads. 

For \textsc{GraphGen}, we use 4 layers of LSTM cells for $f_{trans}$ with hidden state dimension of $256$ and the dimension of $f_{emb}$ is set to $92$. Hidden layers of size $512$ are used in MLPs for $f_{t_u}$, $f_{t_v}$, $f_{L_u}$, $f_{e}$, $f_{L_v}$. We use \emph{Adam optimizer} with a batch size of $32$ for training. We use a dropout of $p=0.2$ in MLP and LSTM layers. Furthermore, we use gradient clipping to remove exploding gradients and L2 regularizer to avoid over-fitting.
 
To evaluate the performance in any dataset, we split it into three parts: the train set, validation set, and test set. Unless specifically mentioned, the default split among training, validation and test is $80\%$, $10\%$, $10\%$ of graphs respectively. We stop the training when validation loss is minimized or less than $0.05\%$ change in validation loss is observed over an extended number of epochs. Note that both Citeseer and Cora represent a single graph. To form the training set in these datasets, we sample subgraphs by performing random walk with restarts from multiple nodes. More specifically, to sample a subgraph, we choose a node with probability proportional to its degree. Next, we initiate random walk with restarts with restart probability $0.15$. The random walks stop after $150$ iterations. Any edge that is sampled at least once during random walk with restarts is part of the sampled subgraph. This process is then repeated $X$ times to form a dataset of $X$ graphs. The sizes of the sampled subgraphs range from $1\leq |V|\leq 102$ and $1\leq |E|\leq 121$ in Citeseer and $9\leq |V|\leq 111$ and $20\leq |E|\leq 124$ in Cora.

\subsubsection{Metrics}
\label{sec:metrics}
The metrics used in the experiments can be classified into the following categories:

\begin{itemize}
    \item \textbf{Structural metrics:} We use the metrics used by GraphRNN\cite{graphrnn} for structural evaluation: {\bf (1)} \emph{node degree} distribution, {\bf (2)} \emph{clustering coefficient} distribution of nodes, and {\bf (3)} \emph{orbit count} distribution, which measures the number of all orbits with $4$ nodes. Orbits capture higher-level motifs that are shared between generated and test graphs. The closer the distributions between the generated and test graphs, the better is the quality. In addition, we also compare the graph size of the generated graphs with the test graphs in terms of {\bf (4)} \emph{Average number of nodes} and {\bf (5)} \emph{Average number of edges.}

    \item \textbf{Label Accounting metrics:} We compare the distribution of {\bf (1)} node labels, {\bf (2)} edge labels, and {\bf (3)} joint distribution of node labels and degree in generated and test graphs.
    
    \item\textbf{Graph Similarity: } To capture the similarity of generated graphs in a more holistic manner that considers both structure and labels, we use \emph{Neighbourhood Sub-graph Pairwise Distance Kernel (NSPDK)}\cite{nspdk}. NSPDK measures the distance between two graphs by matching pairs of subgraphs with different radii and distances. The lower the NSPDK distance, the better is the performance. This quality metric is arguably the most important since it captures the global similarity instead of local individual properties.

    \item \textbf{Redundancy checks:} Consider a generative model that generates the exact same graphs that it saw in the training set. Although this model will perform very well in the above metrics, such a model is practically useless. Ideally, we would want the generated graphs to be diverse and similar, but not identical. To quantify this aspect, we check {\bf (1)} \emph{Novelty}, which measures the percentage of generated graphs that are not subgraphs of the training graphs and vice versa. Note that identical graphs are subgraph isomorphic to each other. In other words, novelty checks if the model has learned to generalize \emph{unseen} graphs. We also compute {\bf(2)} \emph{Uniqueness}, which captures the diversity in generated graphs. To compute \emph{Uniqueness}, we first remove the generated graphs that are subgraph isomorphic to some other generated graphs. The percentage of graphs remaining after this operation is defined as \emph{Uniqueness}. For example, if the model generates $100$ graphs, all of which are identical, the uniqueness is $1/100 = 1\%$.
\end{itemize}

To compute the distance between two distributions, like in Graph\-RNN\cite{graphrnn}, we use \emph{Maximum Mean Discrepancy (MMD)}\cite{metric:mmd}. 
 To quantify quality using a particular metric, we generate $2560$ graphs and compare them with a random sample of $2560$ test graphs. On all datasets except Enzymes, we report the average of $10$ runs of computing metric, comparing $256$ graphs in a single run. Since the Enzymes dataset contains only $575$ graphs, we sample $64$ graphs randomly from the test set $40$ times and report the average.

\subsection{Quality}
\label{sec:quality}
Table~\ref{tab:quality} presents the quality achieved by all three benchmarked techniques across $10$ different metrics on $6$ datasets. We highlight the key observations that emerge from these experiments. Note that, some of the generated graphs may not adhere to the structural properties assumed in \S~\ref{sec:formulation} \textit{i.e.}, no self loops, multiples edges or disconnected components, so we prune all the self edges and take the maximum connected component for each generated graph.

\textbf{Graph and Sub-graph level similarity: } On the NSPDK metric, which is the most important due to capturing the global similarity of generated graphs with test graphs, \textsc{GraphGen} is significantly better across all datasets. This same trend also transfers to orbit count distributions. Orbit count captures the presence of motifs also seen in the test set. These two metrics combined clearly establishes that \textsc{GraphGen} model graphs better than GraphRNN and DeepGMG.

\textbf{Node-level metrics: }Among the other metrics, \textsc{GraphGen} remains the dominant performer. To be more precise, \textsc{GraphGen} is consistently the best in both Node and Edge Label preservation, as well as graph size in terms of the number of edges. It even performs the best on the joint Node Label and Degree metric, indicating its superiority in capturing structural and semantic information together. The performance of \textsc{GraphGen} is comparatively less impressive in the Enzymes dataset, where GraphRNN marginally outperforms in the Degree and Clustering Coefficient metrics. Nonetheless, even in Enzyme, among the eight metrics, \textsc{GraphGen} is better in six. This result also highlights the need to not rely on only node-level metrics. Specifically, although GraphRNN models the node degrees and clustering coefficients well, it generates graphs that are much smaller in size and hence the other metrics, including NSPDK, suffers.

\textbf{Novelty and Uniqueness: }Across all datasets, \textsc{GraphGen} has uniqueness of at least $98\%$, which means it does not generate the same graph multiple times. In contrast, GraphRNN has a significantly lower uniqueness in chemical compound datasets. In Lung, only $45\%$ of the generated graphs are unique. In several of the datasets, GraphRNN also has low novelty, indicating it regenerates graphs (or subgraphs) it saw during training. For example, in Cora, Citeseer and, the mixed chemical dataset, at least $30\%$ of the generated graphs are regenerated from the training set. While we cannot pinpoint the reason behind this performance, training on random graph sequence representations could be a factor. More specifically, even though the generative model may generate new sequences, they may correspond to the same graph. While this is possible in \textsc{GraphGen} as well, the likelihood is much less as it is trained on DFS codes that enable one-to-one mapping with graphs. 


\textbf{Analysis of GraphRNN and DeepGMG: }Among GraphRNN and DeepGMG, DeepGMG generally performs better in most metrics. However, DeepGMG is extremely slow and fails to scale on larger networks due to 
 $O(|V||E|^2)$ complexity. 

GraphRNN's major weakness is in learning graph sizes. As visible in Table~\ref{tab:quality}, GraphRNN consistently generates graphs that are much smaller than the test graphs. We also observe that this issue arises only in labeled graphs. If the labels are removed while retaining their structures, GraphRNN generates graphs that correctly mirror the test graph size distribution. This result clearly highlights that labels introduce an additional layer of complexity that simple extensions of models built for unlabeled graphs do not solve.

\subsubsection{Visual Inspection}
\begin{figure}[t]
\centering
\subfigure[Real graphs]{
\includegraphics[width=1.6in]{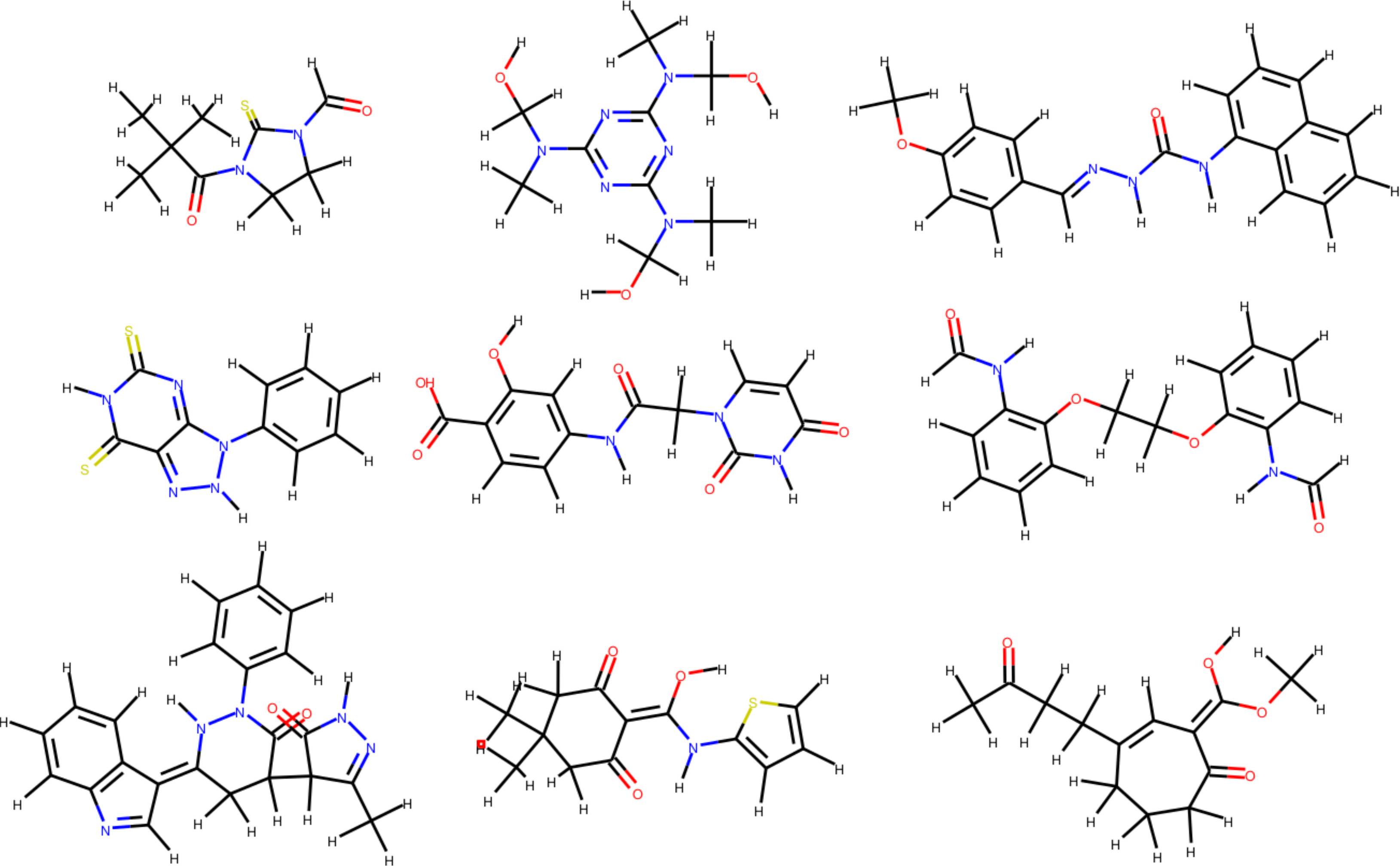}}
\subfigure[GraphGen]{
\includegraphics[width=1.6in]{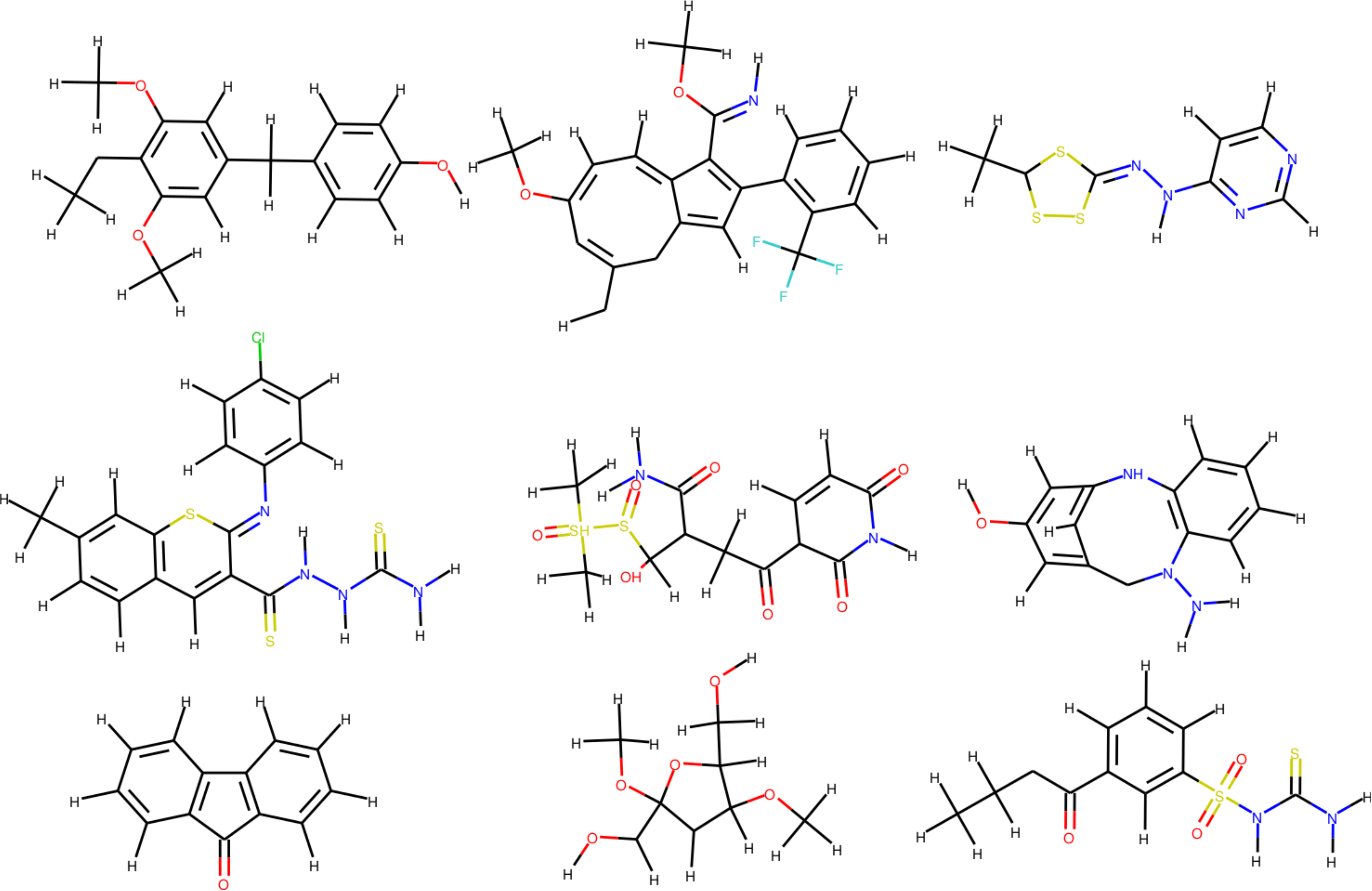}}
\subfigure[GraphRNN]{
\includegraphics[width=1.6in]{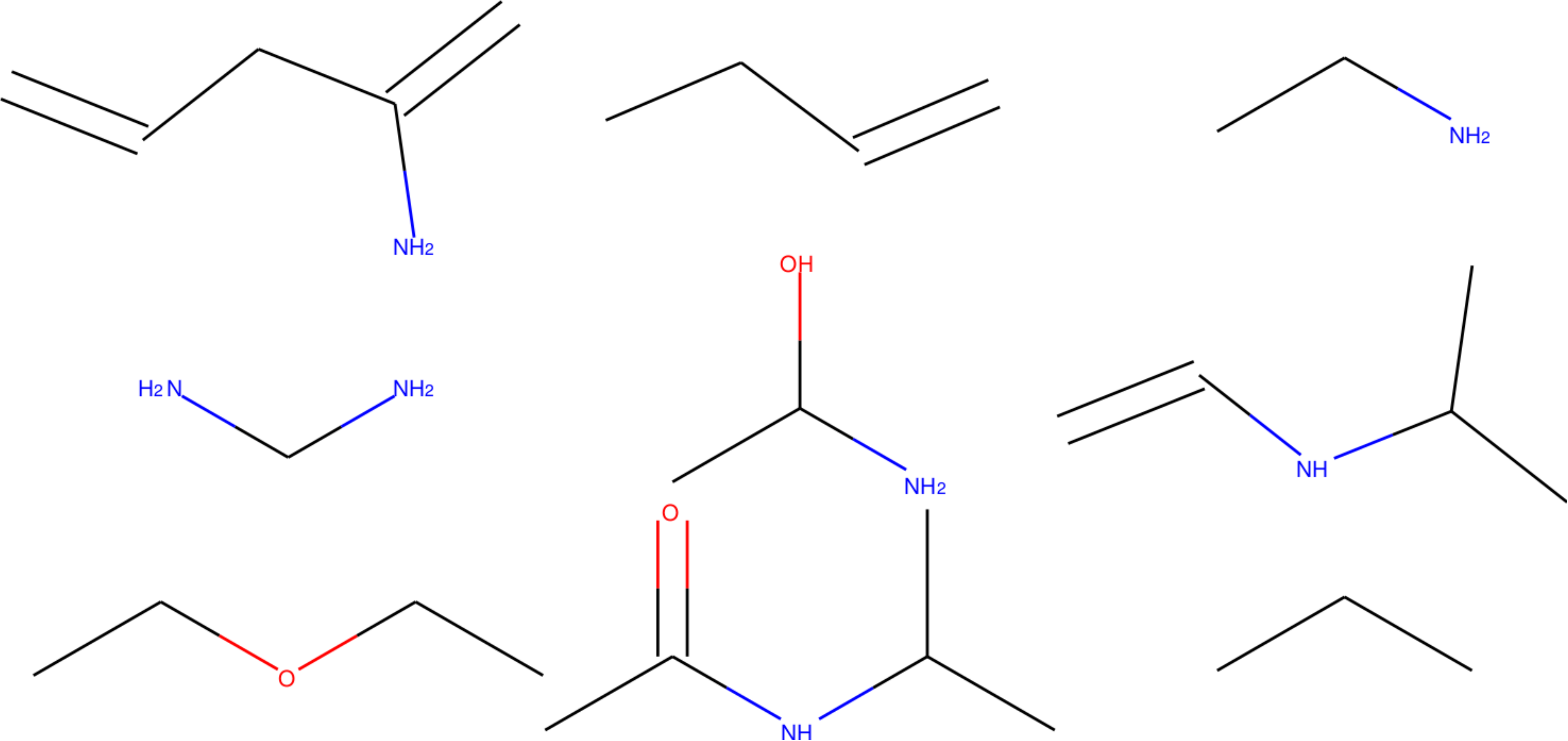}}
\subfigure[DeepGMG]{
\includegraphics[width=1.6in]{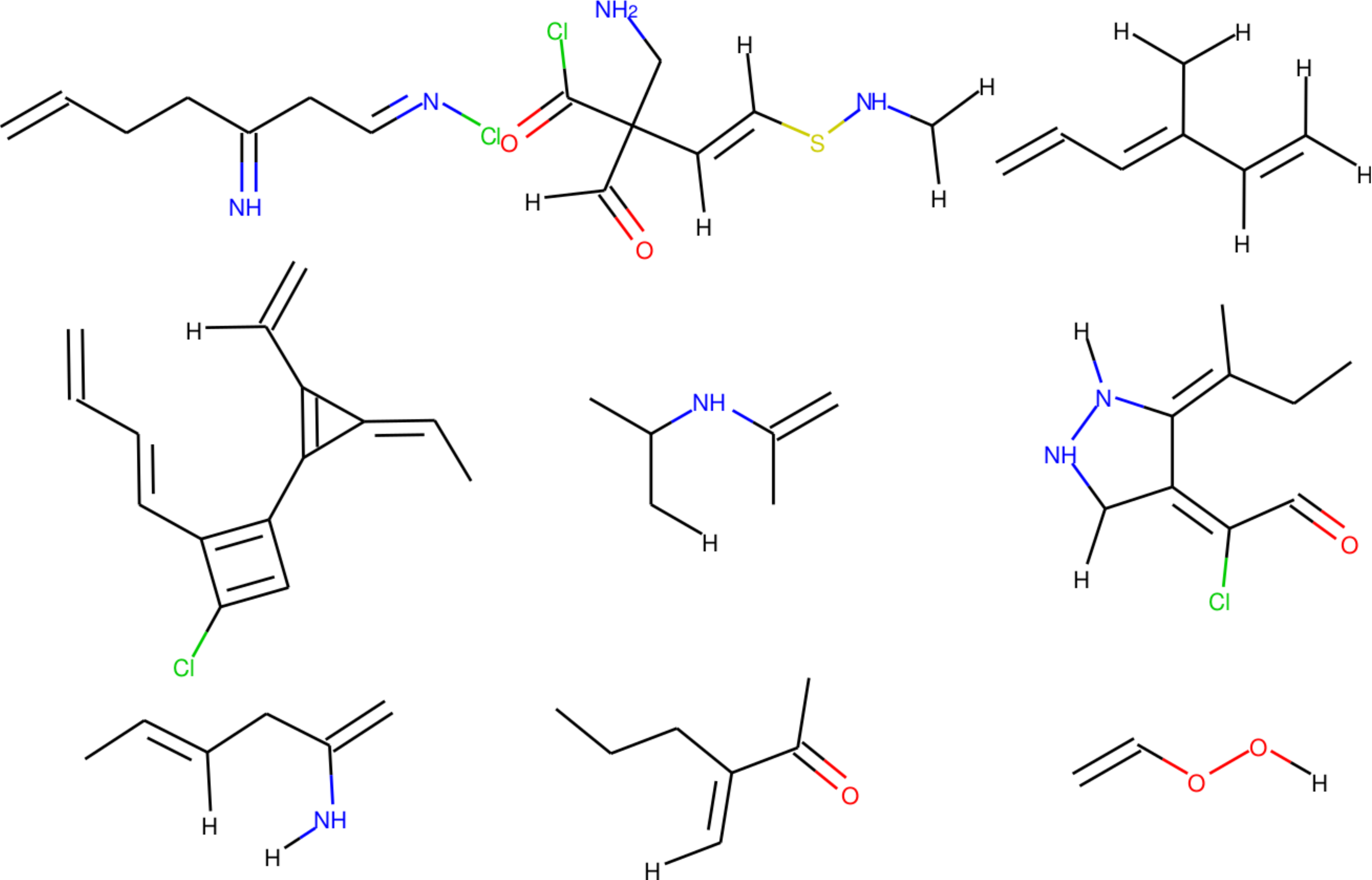}}
\caption { Visual comparison of the generated graphs with real graphs from the Lung dataset.
}
\label{fig:chemical}
\end{figure}

To gain visual feedback on the quality of the generated graphs, in Fig.~\ref{fig:chemical}, we present a random sample of $9$ graphs from the training set of Lung dataset and those generated by each of the techniques. Visually, \textsc{GraphGen} looks the most similar, while GraphRNN is the most dissimilar. This result is consistent with the quantitative results obtained in Table~\ref{tab:quality}. For example, consistent with our earlier observations, the GraphRNN graphs are much smaller in size and lack larger motifs like benzene rings. In contrast, \textsc{GraphGen}'s graphs are of similar sizes and structures.

\begin{figure*}[t]
\centering
\subfigure[Real]{
\includegraphics[width=2.0in]{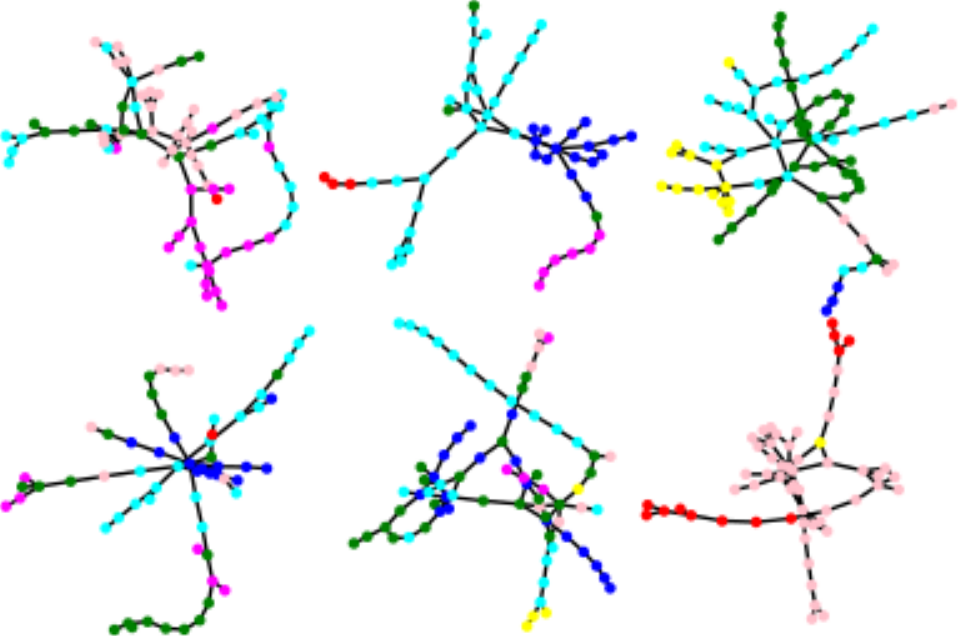}}
\hfill
\subfigure[GraphGen]{
\label{fig:nonchemical_graphgen}
\includegraphics[width=2.0in]{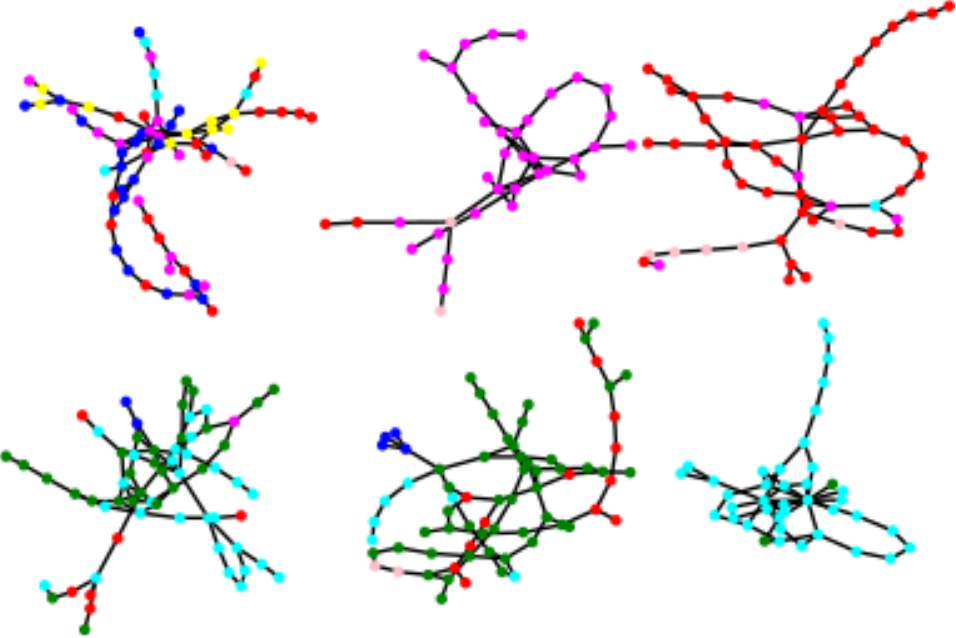}}
\hfill
\subfigure[GraphRNN]{
\includegraphics[width=2.0in]{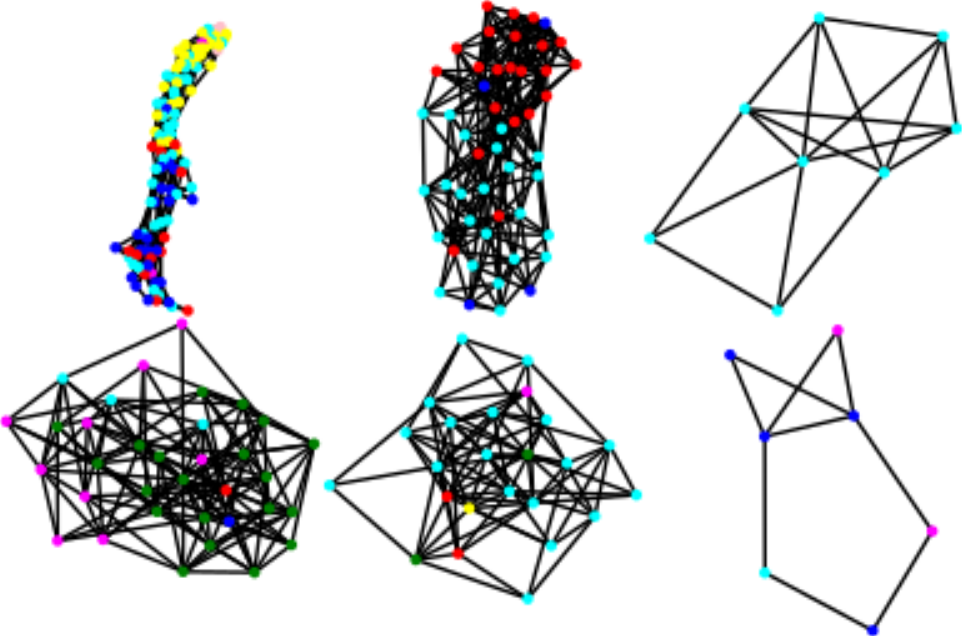}}
\vspace{-0.1in}
\caption { Visual comparison of the generated graphs with real graphs from Cora. In this figure, the colors of the nodes denote their labels. 
}
\label{fig:nonchemical}
\end{figure*}

In Fig.~\ref{fig:nonchemical}, we perform the same exercise as above with $6$ randomly picked graphs from the Cora dataset and those generated by \textsc{GraphGen} and GraphRNN. Since the graphs in this dataset are much larger, it is hard to comment if \textsc{GraphGen} is better than GraphRNN through visual inspection; the layout of a graph may bias our minds. However, we notice one aspect where \textsc{GraphGen} performs well. Since Cora is a citation network, densely connected nodes (communities) have a high affinity towards working in the same publication area (node label/color). This correlation of label homogeneity with communities is also visible in \textsc{GraphGen}(Fig.~\ref{fig:nonchemical_graphgen}).

\subsection{Scalability}
\label{sec:scalability}
After establishing the clear superiority of \textsc{GraphGen} in quality for labeled graph generative modeling, we turn our focus to scalability. To enable modeling of large graph databases, the training must complete within a reasonable time span. The penultimate column in Table~\ref{tab:quality} sheds light on the scalability of the three techniques. As clearly visible, \textsc{GraphGen} is $2$ to $5$ times faster than GraphRNN, depending on the dataset. DeepGMG is clearly the slowest, which is consistent with its theoretical complexity of $O(|V||E|^2)$. If we exclude the sequence generation aspect, GraphRNN has a lower time complexity for training than \textsc{GraphGen}; while GraphRNN is linear to the number of nodes in the graph, \textsc{GraphGen} is linear to the number of edges. However, \textsc{GraphGen} is still able to achieve better performance due to training on canonical labels of graphs.  In contrast, GraphRNN generates a random sequence representation for a graph in each epoch. Consequently, as shown in the last column of Table~\ref{tab:quality}, GraphRNN runs for a far larger number of epochs to achieve loss minimization in the validation set. Note that DeepGMG converges within the minimum number of epochs. However, the time per epoch of DeepGMG is $100$ to $200$ times higher than \textsc{GraphGen}. Compared to GraphRNN, the time per epoch of \textsc{GraphGen} is $\approx 30\%$ faster on average.

\subsubsection{Impact of Training Set Size} To gain a deeper understanding of the scalability of \textsc{GraphGen}, we next measure the impact of training set size on performance. Figs.~\ref{fig:sizevstime}-\ref{fig:sizevstime_cora} present the growth of training times on ZINC and Cora against the number of graphs in the training set. While ZINC contains 3.2 million graphs, in Cora, we sample up to 1.2 million subgraphs from the original network for the training set. As visible, both GraphRNN and DeepGMG has a much steeper growth rate than \textsc{GraphGen}. In fact, both these techniques fail to finish within 3 days for datasets exceeding 100,000 graphs. In contrast, \textsc{GraphGen} finishes within two and a half days, even on a training set exceeding $3$ million graphs. The training times of all techniques are higher in Cora since the graphs in this dataset are much larger in size. We also observe that the growth rate of \textsc{GraphGen} slows at larger training sizes. On investigating further, we observe that with higher training set sizes, the number of epochs required to reach the validation loss minima reduces. Consequently, we observe the trend visible in Figs.~\ref{fig:sizevstime}-\ref{fig:sizevstime_cora}. This result is not surprising since the learning per epoch is higher with larger training sets.

An obvious question arises at this juncture: \textit{Does lack of scalability to larger training sets hurt the quality of GraphRNN and DeepGMG?} Figs.~\ref{fig:sizevsnspdk}-\ref{fig:sizevsorbit_cora} answer this question. In both ZINC and Cora, we see a steep improvement in the quality (reduction in NSPDK and Orbit) of GraphRNN as the training size is increased from $\approx 10,000$ graphs to $\approx 50,000$ graphs. The improvement rate of \textsc{GraphGen} is relatively milder, which indicates that GraphRNN has a greater need for larger training sets. However, this requirement is not met as GraphRNN fails to finish within $3$ days, even for $100,000$ graphs. Overall, this result establishes that scalability is not only important for higher productivity but also improved modeling.

\subsubsection{Impact of Graph Size}
\label{sec:impact_graphsize}
\begin{figure*}[t]
\centering

\subfigure[ZINC]{
\label{fig:sizevstime}
\includegraphics[width=1.30in]{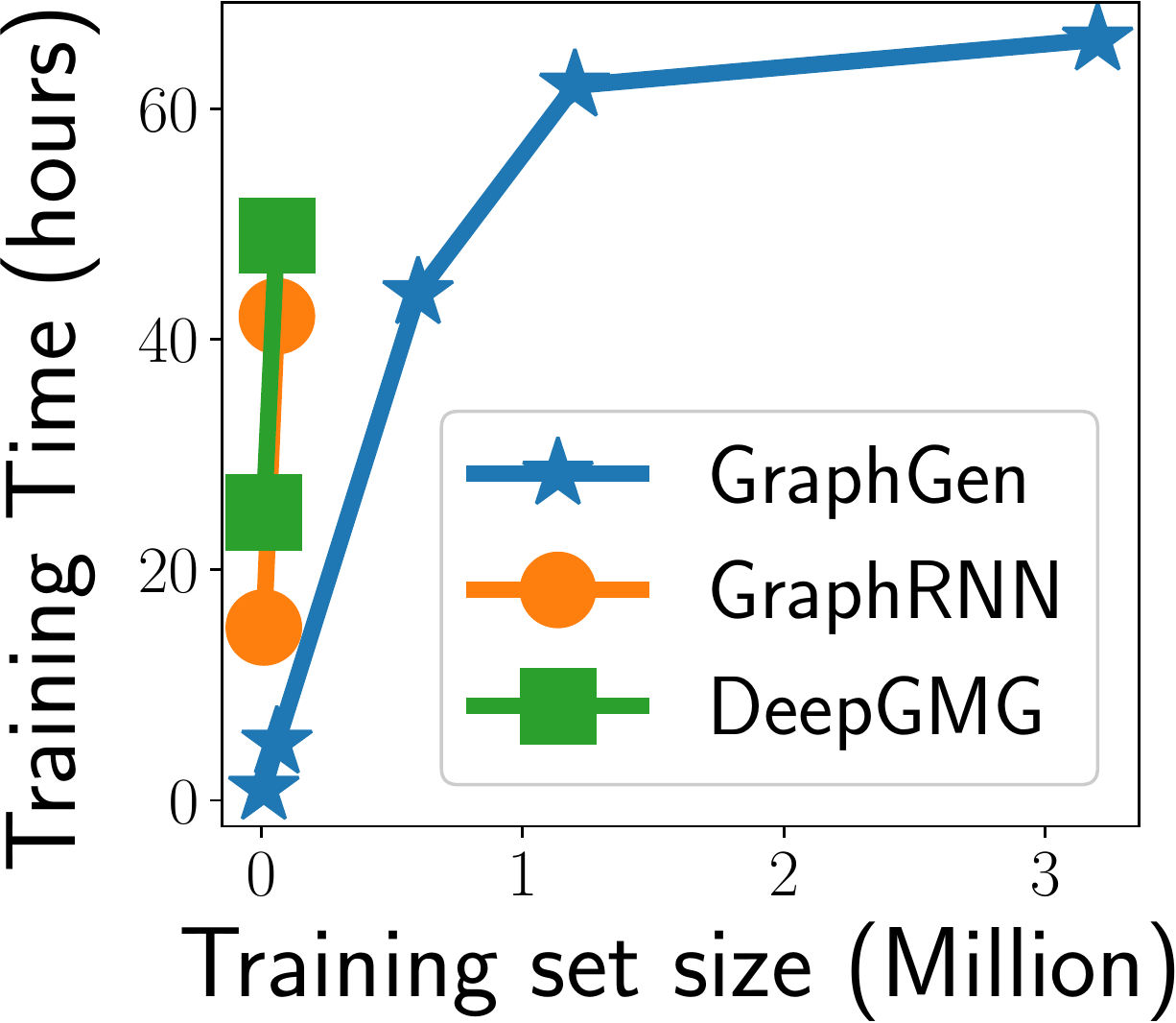}
}
\subfigure[Cora]{
\label{fig:sizevstime_cora}
\includegraphics[width=1.30in]{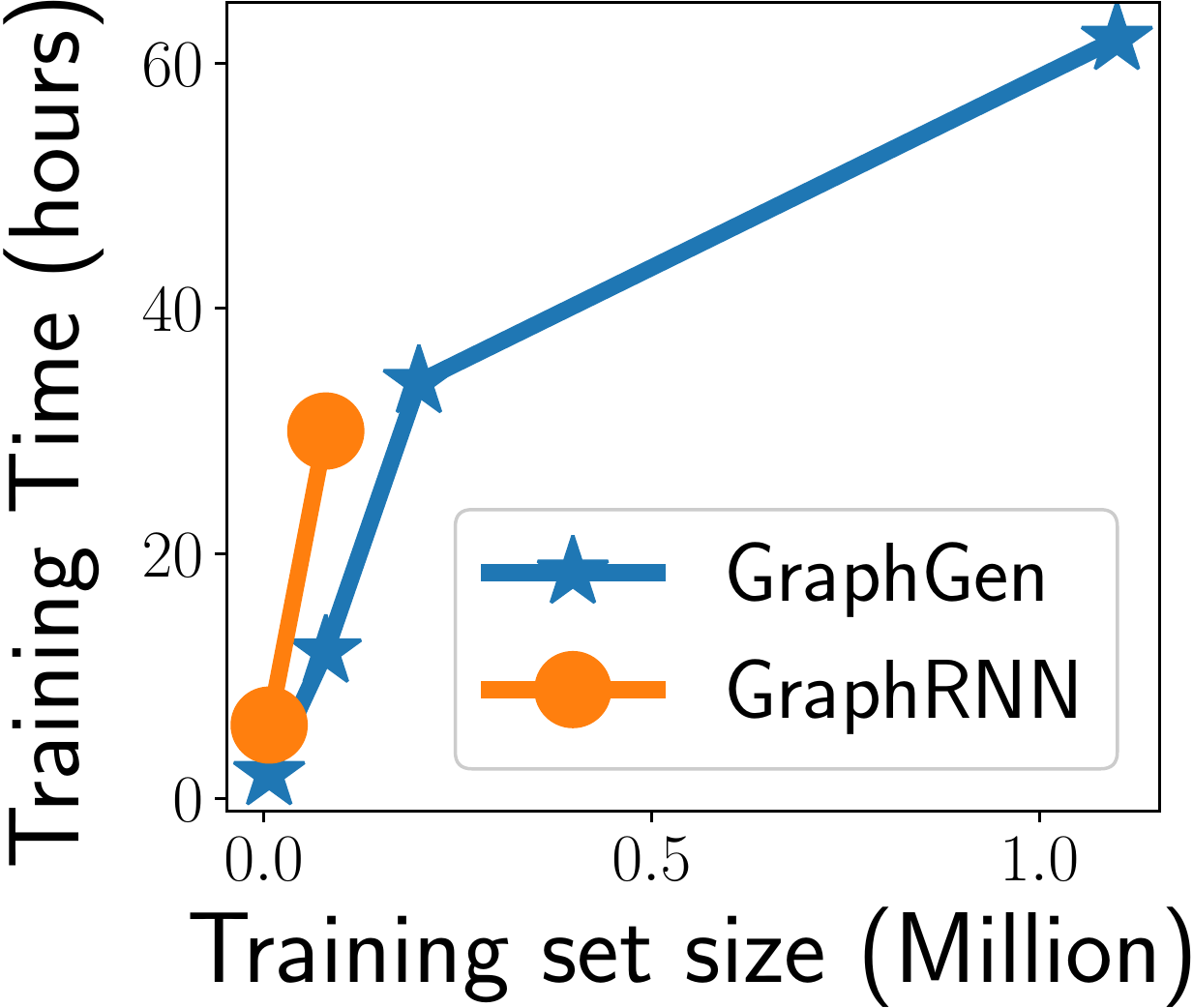}
}
\subfigure[ZINC]{
\label{fig:sizevsnspdk}
\includegraphics[width=1.30in]{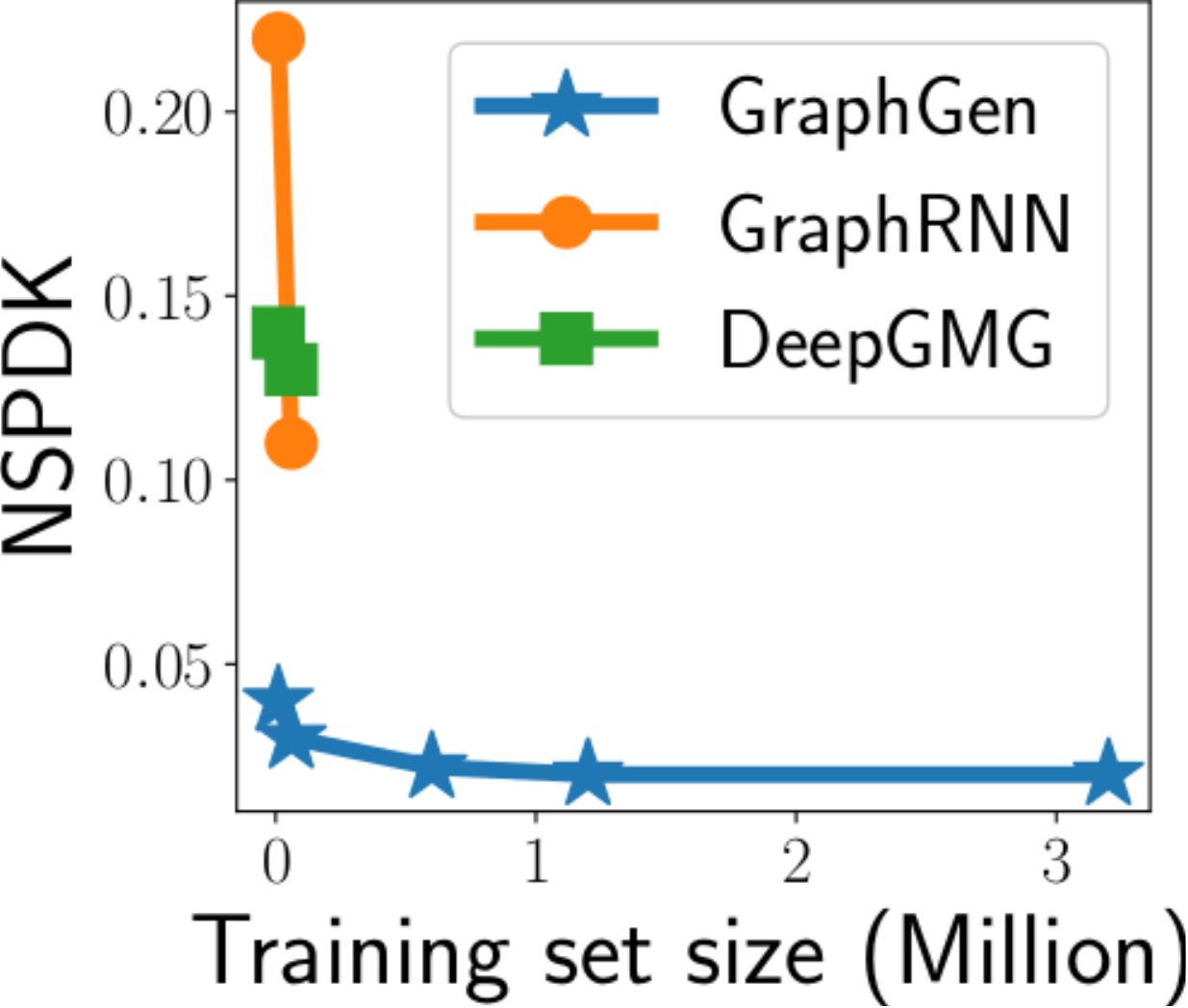}
}
\subfigure[ZINC]{
\label{fig:sizevsorbit}
\includegraphics[width=1.30in]{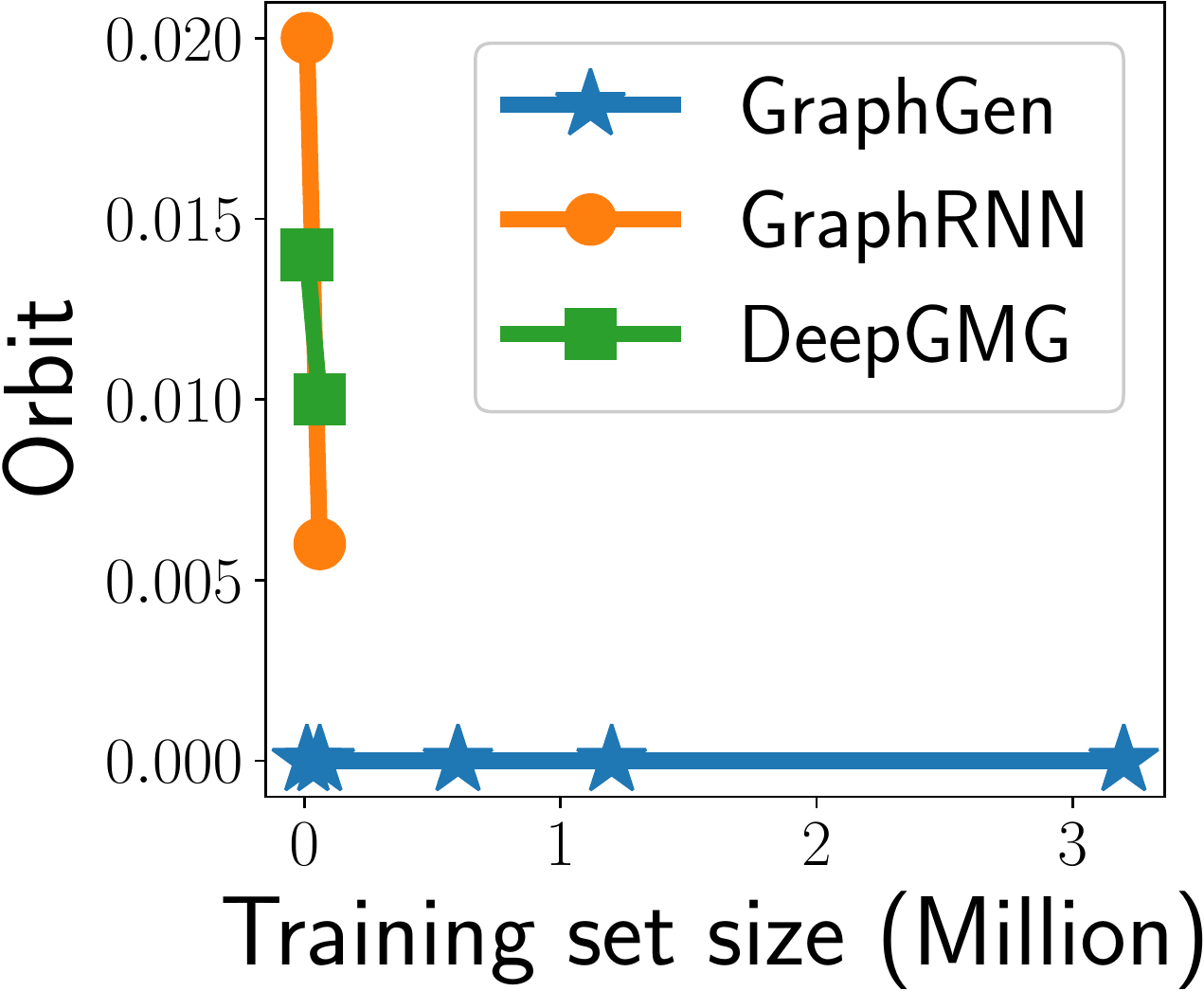}
}
\subfigure[Cora]{
\label{fig:sizevsorbit_cora}
\includegraphics[width=1.30in]{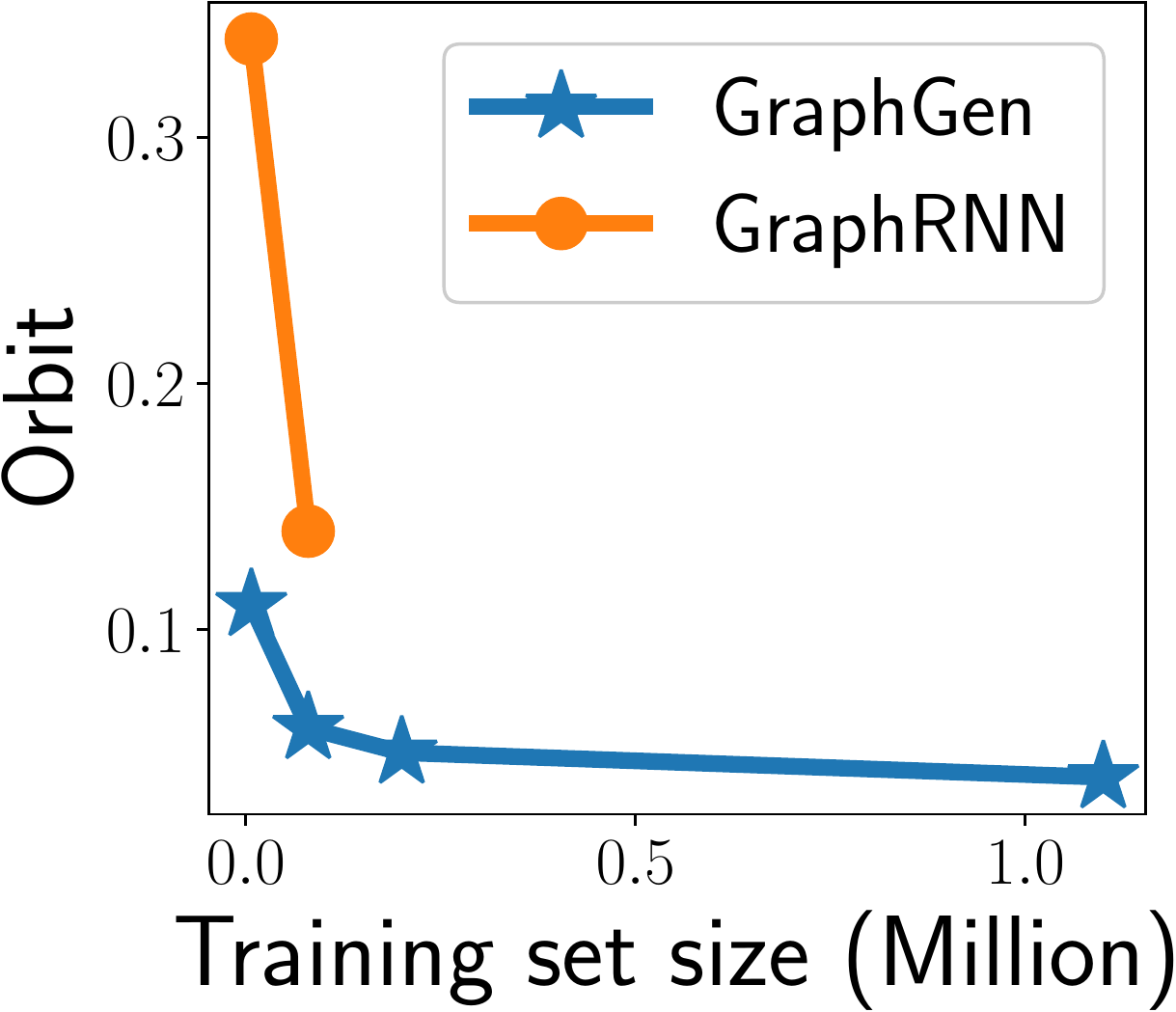}
}\\
\subfigure[All chemical]{
\label{fig:graphvsnspdk_chem}
\includegraphics[width=1.60in]{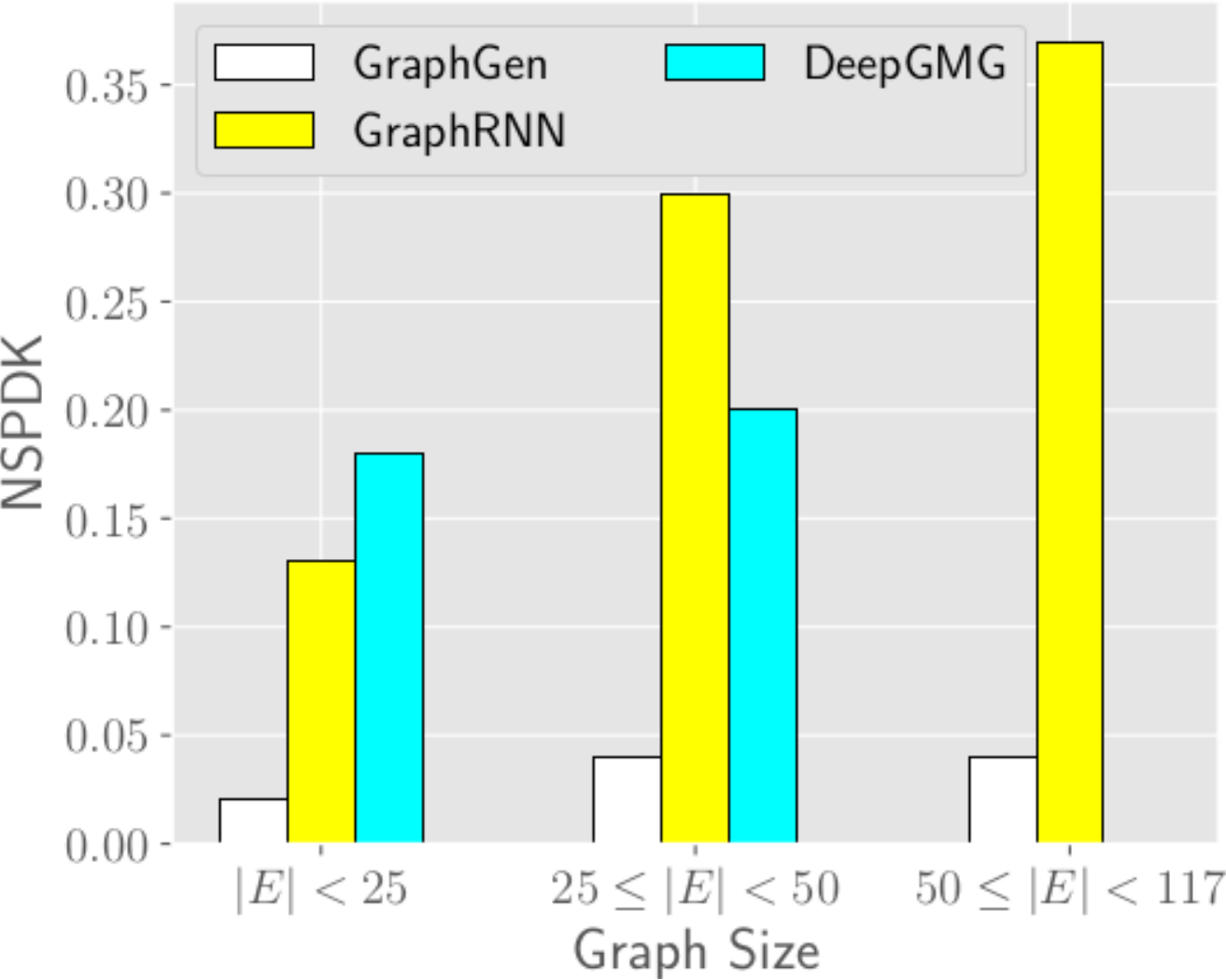}
}
\subfigure[All chemical]{
\label{fig:graphvssize_chem}
\includegraphics[width=1.5in]{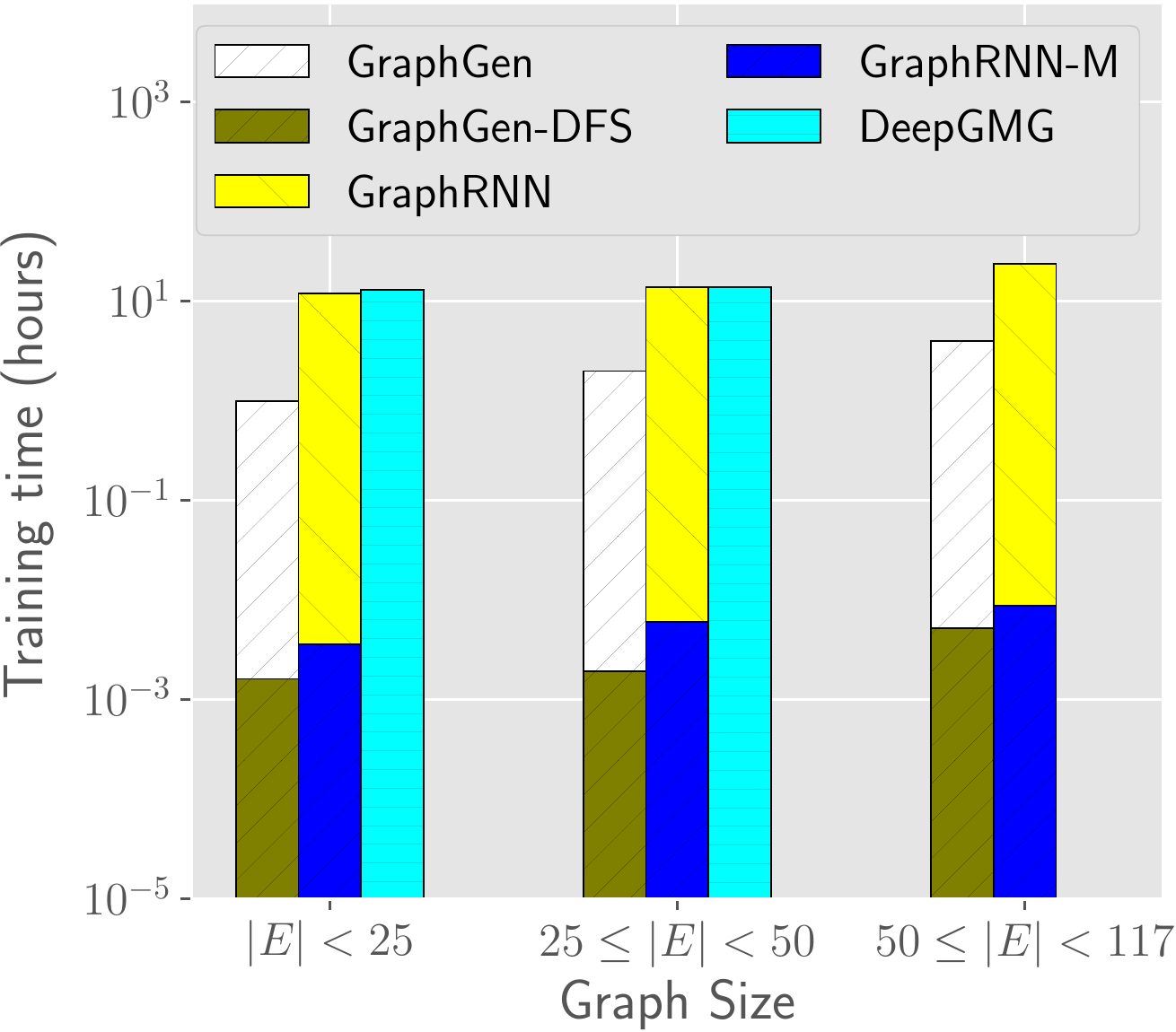}
}
\subfigure[Cora]{
\label{fig:graphvsnspdk_cora}
\includegraphics[width=1.65in]{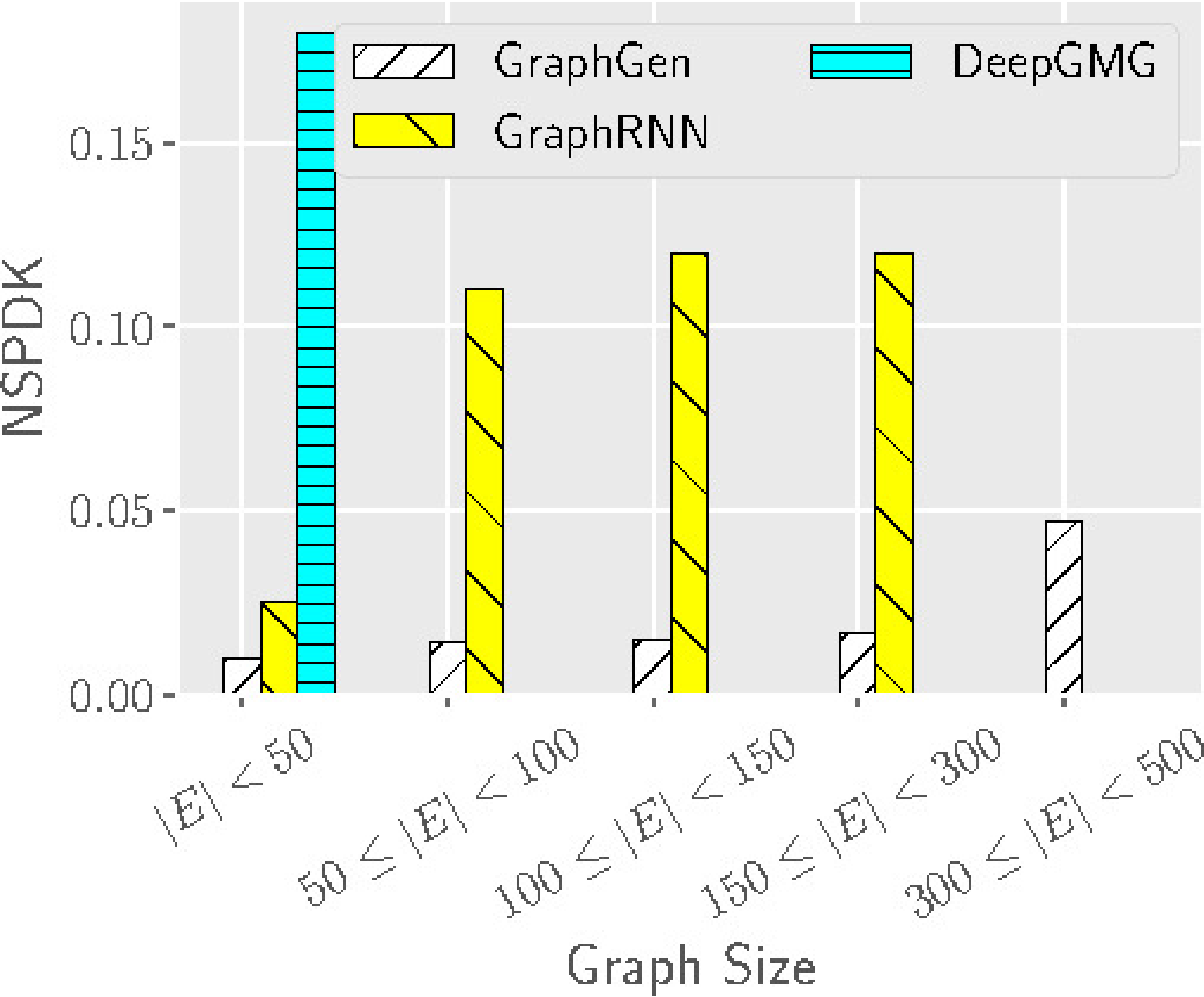}
}
\subfigure[Cora]{
\label{fig:graphvssize_cora}
\includegraphics[width=1.7in]{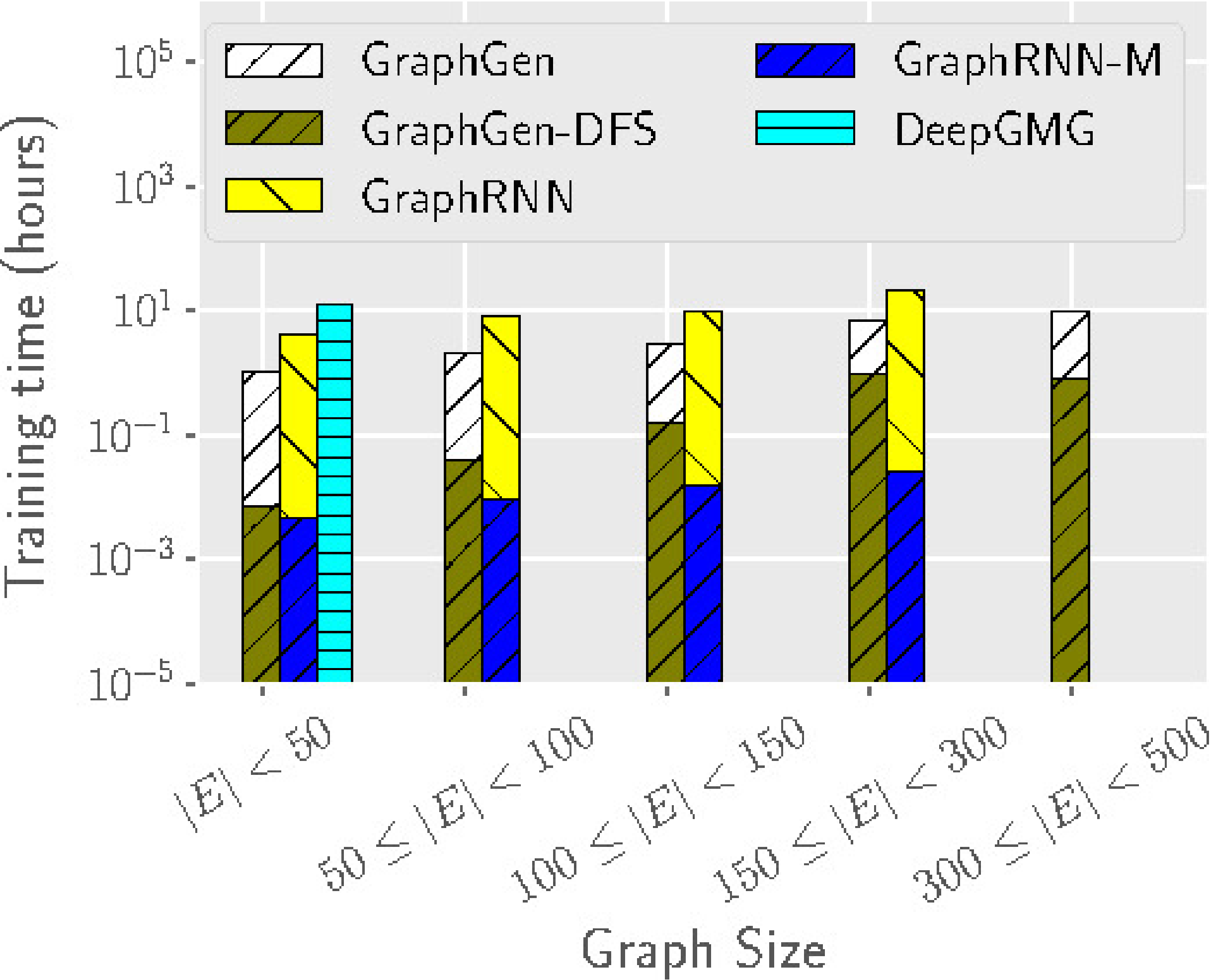}
}
\vspace{-0.1in}
\caption{Impact of training set size on (a-b) efficiency and (c-e) quality. Impact of graph size on (f, h) quality and (g, i) training time. The $y$-axis of (g) and (i) are in log scale.}
\label{fig:trainingsize}
\end{figure*}

We next evaluate how the size of the graph itself affects the quality and training time. Towards that end, we partition graphs in a dataset into multiple buckets based on its size. Next, we train and test on each of these buckets individually and measure the impact on quality and efficiency. Figs.\ref{fig:graphvsnspdk_chem} and \ref{fig:graphvsnspdk_cora} present the impact on quality in chemical compounds (same as Mixed dataset in Table~\ref{tab:quality}) and Cora respectively. In both datasets, each bucket contains exactly $10,000$ graphs. The size of graphs in each bucket is however different, as shown in the $x$-axis of Figs.\ref{fig:graphvsnspdk_chem} and \ref{fig:graphvsnspdk_cora}. Note that Cora contains larger graphs since the network itself is much larger than chemical compounds.  The result for DeepGMG is missing on the larger graph-size buckets since it fails to model large graphs. GraphRNN runs out of GPU memory ($12 GB$) in the largest bucket size in Cora.

As visible, there is a clear drop in quality (increase in NSPDK) as the graph sizes grow. This result is expected since larger graphs involve larger output spaces, and the larger the output space, the more complex is the modeling task. Furthermore, both \textsc{GraphGen} and GraphRNN convert graph modeling into sequence modeling. It is well known from the literature that auto-regressive neural models struggle to learn \emph{long-term} dependencies in sequences. When the sequence lengths are large, this limitation impacts the quality. Nonetheless, \textsc{GraphGen} has a more gradual decrease in quality compared to GraphRNN and DeepGMG.

\begin{figure}[b]
\centering
\subfigure[Quality]{
\label{fig:invariantvsquality}
\includegraphics[width=1.5in]{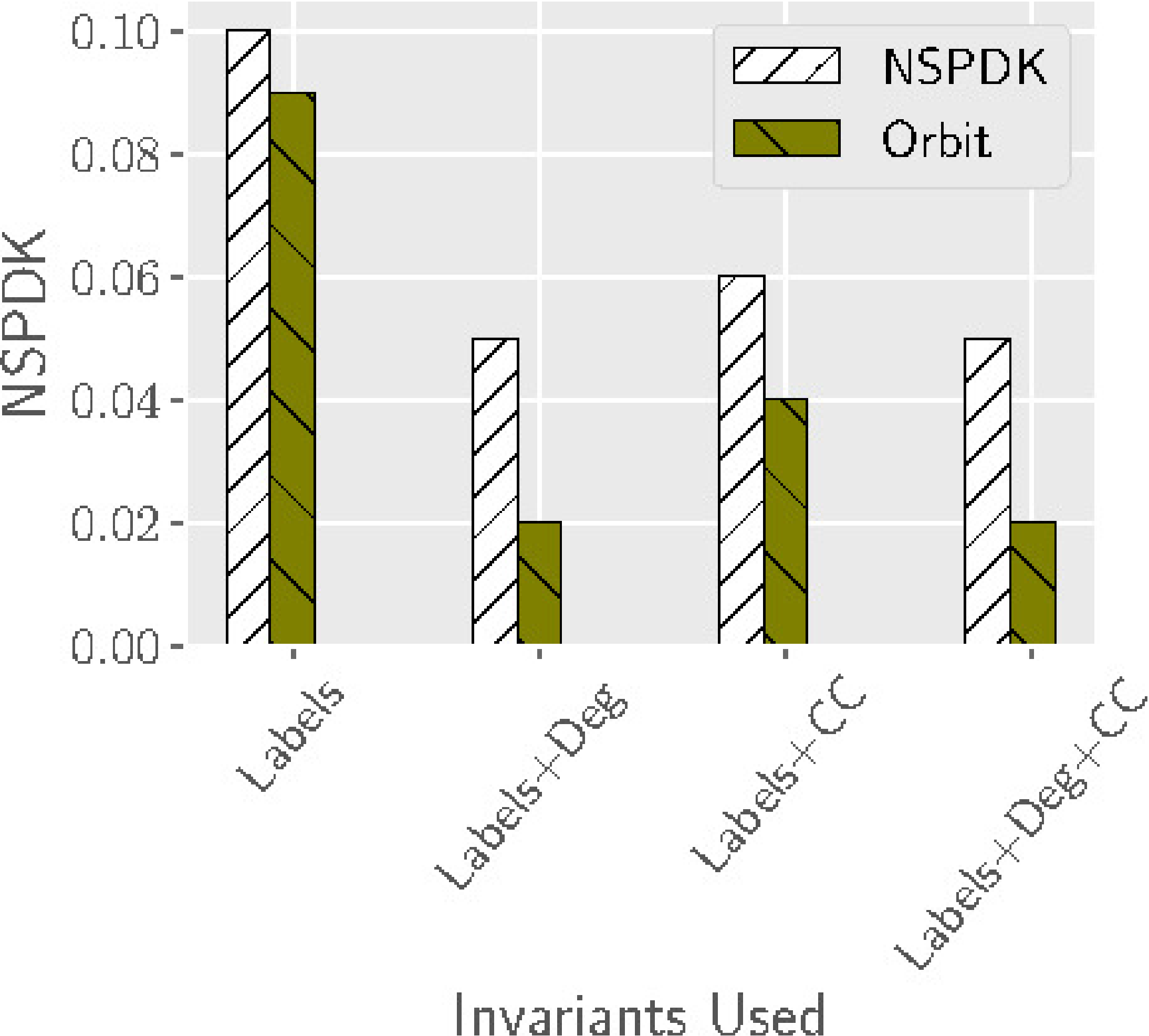}
}
\subfigure[Efficiency]{
\label{fig:invariantvstime}
\includegraphics[width=1.5in]{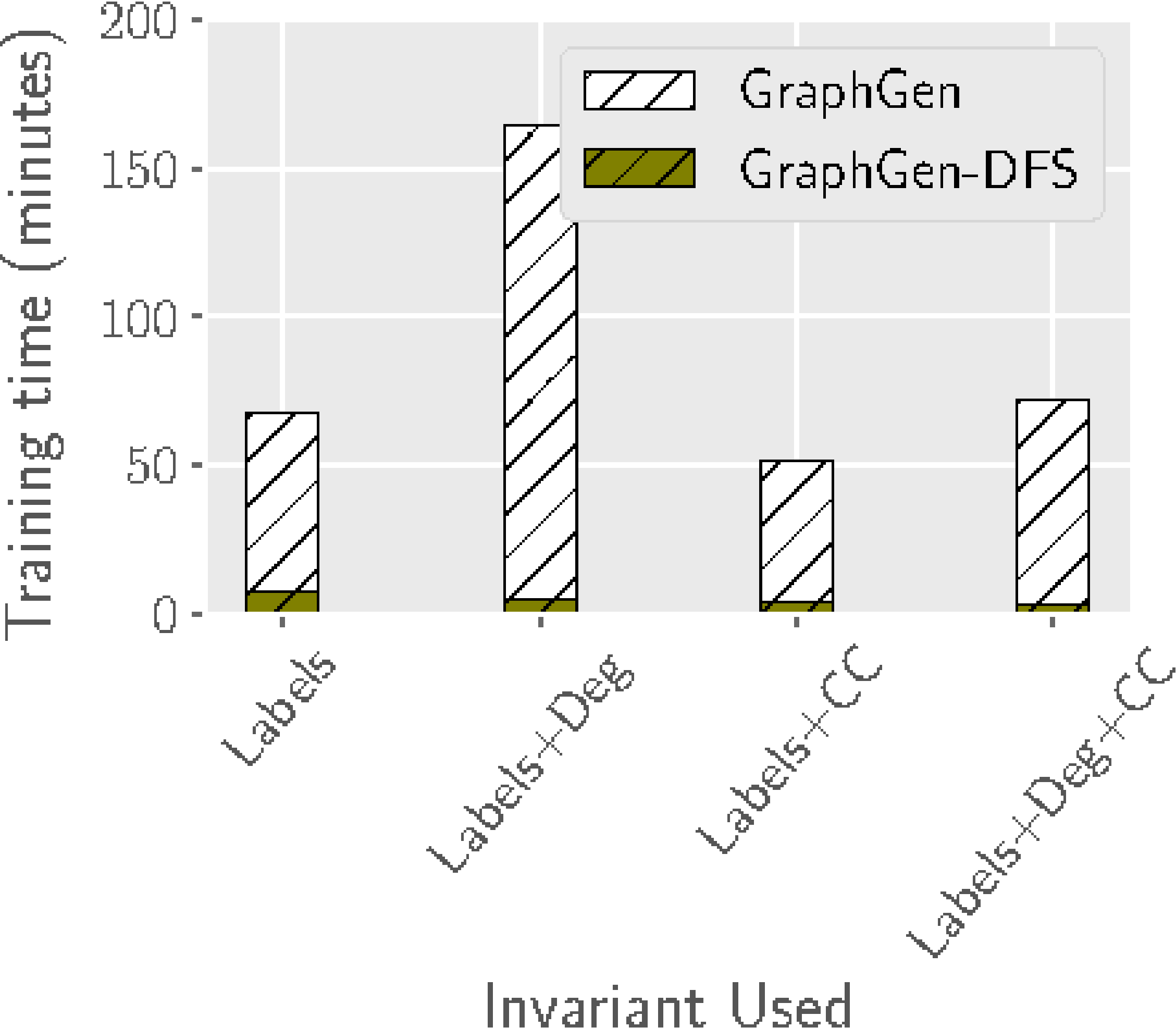}
}

\vspace{-0.1in}
\caption{Impact of vertex invariants on (a) quality and (b) training time on Enzymes dataset}
\label{fig:invariants}
\end{figure}

Next, we analyze the impact of graph size on the training time. Theoretically, we expect the training time to grow, and this pattern is reflected in Figs.~\ref{fig:graphvssize_chem} and \ref{fig:graphvssize_cora}. Note that the $y$-axis is in log scale in these plots. Consistent with previous results, \textsc{GraphGen} is faster. In addition to the overall running time, we also show the average time taken by \textsc{GraphGen} in computing the minimum DFS code. Recall that the worst-case complexity of this task is factorial to the number of nodes in the graph. However, we earlier argued that in practice, the running times are much smaller. This experiment adds credence to this claim by showing that DFS code generation is a small portion of the total running time. For example, in the largest size bucket of chemical compounds, DFS code generation of all $10,000$ graphs take $30$ seconds out of the total training time of $4$ hours. In Cora, on the largest bucket size of $500$ edges, the DFS construction component consumes $50$ minutes out of the total training time of $10$ hours. GraphRNN has a similar pre-processing component as well (GraphRNN-M), where it generates a large number of BFS sequences to estimate a hyper-parameter. We find that this pre-processing time, although of polynomial-time complexity, is slower than minimum DFS code computation in chemical compounds. Overall, the results of this experiment support the proposed strategy of feeding canonical labels to the neural model instead of random sequence representations.

\subsection{Impact of Invariants}
\label{sec:invariants}

Minimum DFS code generation relies on the node and edge labels to prune the search space and identify the lexicographically smallest code quickly. When graphs are unlabeled or have very few labels, this process may become more expensive. In such situations, labels based on vertex invariants allow us to augment existing labels. To showcase this aspect, we use the Enzymes dataset, which has only $3$ node labels and no edge labels. We study the training time and the impact on quality based on the vertex invariants used to label nodes. We study four combinations: {\bf (1)} existing node labels only, {\bf (2)} node label + node degree, {\bf (3)} node label + clustering coefficient \emph{(CC)}, and {\bf (4)} label + degree + CC. Fig.~\ref{fig:invariants} presents the results. As visible, there is a steady decrease in the DFS code generation time as more invariants are used. No such trend, however, is visible in the total training time as additional features may both increase or decrease the time taken by the optimizer to find the validation loss minima. From the quality perspective, both NSPDK and Orbit show improvement (reduction) through the use of invariants. Between degree and CC, degree provides slightly improved results. Overall, this experiment shows that invariants are useful in both improving the quality and reducing the cost of minimum DFS code generation.

\subsection{Alternatives to LSTMs}
For deep, auto-regressive modeling of sequences, we design an LSTM tailored for DFS codes. \textit{Can other neural architectures be used instead? Since a large volume of work has been done in the NLP community on sentence modeling, can those algorithms be used by treating each edge tuple as a word?} To answer these questions, we replace LSTM with SentenceVAE\cite{sentencevae} and measure the performance. We present the results only in Lung and Cora due to space constraints; a similar trend is observed in other datasets as well. In Table~\ref{tab:sentence}, we present the two metrics that best summarizes the impact of this experiment. As clearly visible, SentenceVAE introduces a huge increase in NSPDK distance, which means inferior quality. An even more significant result is observed in Uniqueness, which is close to $0$ for SentenceVAE. This means the same structure is repeatedly generated by SentenceVAE. In DFS codes, every tuple is unique due to the timestamps, and hence sentence modeling techniques that rely on repetition of words and phrases, fail to perform well. While this result does not say that LSTMs are the only choice, it does indicate the need for architectures that are custom-made for DFS codes.

\begin{table}[b]
\centering
\caption{LSTM Vs. SentenceVAE. }
\vspace{-0.1in}
\label{tab:sentence}
{\scriptsize
\begin{tabular}{|l|c|c|c|c|}
\hline
Technique & \multicolumn{2}{c|}{Lung} &\multicolumn{2}{c|}{Cora}\\
\hline
&NSPDK & Uniqueness & NSPDK & Uniqueness\\
\hline
\textsc{LSTM} &$\mathbf{0.03}$  & $\mathbf{\approx 100}\%$ & $\mathbf{0.012}$ & $\mathbf{98\%}$ \\
\hline
SentenceVAE & $0.7$ & $0.4\%$ & $0.78$ & $0.4\%$  \\
\hline
\end{tabular}}
\end{table}

\section{Concluding Insights}
In this paper, we studied the problem of learning generative models for labeled graphs in a domain-agnostic and scalable manner. There are two key takeaways from the conducted study. First, existing techniques model graphs either through their adjacency matrices or sequence representations. In both cases, the mapping from a graph to its representation is one-to-many. Existing models feed multiple representations of the same graph to the model and rely on the model overcoming this information redundancy. In contrast, we construct canonical labels of graphs in the form of minimum DFS codes and reduce the learning complexity. Canonical label construction has non-polynomial time complexity, which could be a factor behind this approach not being adequately explored. Our study shows that although the worst-case complexity is factorial to the number of nodes, by exploiting nodes/edge labels, and vertex invariants, DFS-code construction is a small component of the overall training time (Figs.~\ref{fig:graphvssize_chem} and \ref{fig:graphvssize_cora}). Consequently, the time complexity of graph canonization is not a practical concern and, feeding more precise graph representations to the training model is a better approach in terms of quality and scalability. 

The second takeaway is the importance of quality metrics. Evaluating graph quality is a complicated task due to the multi-faceted nature of labeled structures. It is, therefore, important to deploy enough metrics that cover all aspects of graphs such as local node-level properties, structural properties in the form of motifs and global similarity, graph sizes, node and edge labels, and redundancy in the generated graphs. A learned model is effective only if it performs well across all of the metrics. As shown in this study, \textsc{GraphGen} satisfies this criteria.

The conducted study does not solve all problems related to graph modeling. Several graphs today are annotated with feature vectors. Examples include property graphs and knowledge graphs. Can we extend label modeling to feature vectors? All of the existing learning-based techniques, including \textsc{GraphGen}, cannot scale to graphs containing millions of nodes. Can we overcome this limitation? We plan to explore these questions further by extending the platform provided by \textsc{GraphGen}. 

\begin{acks}
The authors thank Lovish Madaan for helpful discussions and comments on the paper. The authors also thank IIT Delhi HPC facility for the computational resources.
\end{acks}

\newpage

\bibliographystyle{ACM-Reference-Format}
\bibliography{main}

\end{document}